\documentclass{article}

\usepackage[final]{neurips_2024}

\usepackage[utf8]{inputenc} %
\usepackage[T1]{fontenc}    %
\usepackage{hyperref}       %
\usepackage{url}            %
\usepackage{booktabs}       %
\usepackage{amsfonts}       %
\usepackage{nicefrac}       %
\usepackage{microtype}      %
\usepackage{xcolor}         %

\usepackage{amsmath,amssymb}

\usepackage{iftex}
\usepackage{siunitx}
\usepackage{wrapfig}
\usepackage{array}
\usepackage{multirow}
\usepackage{tikz}
\usepackage{amsmath}
\usepackage{amsfonts}
\usepackage{amssymb}
\usepackage{capt-of}
\usepackage{booktabs}

\newcommand{\topline}{\toprule} %
\newcommand{\midline}{\midrule} %
\newcommand{\botline}{\bottomrule} %

\setcitestyle{authoryear,open={(},close={)}} %
\usetikzlibrary{positioning, shapes.geometric, arrows.meta}

\usepackage{subfiles}

\newcommand{\hresvthree}[0]{ORCAst}

\newcommand{\phaseOne}[0]{stage 1}

\newcommand{\phaseTwo}[0]{stage 2}
\newcommand{\phaseThree}[0]{stage 3}

\newcommand{\metricdesc}[0]{See Section~\ref{method:metrics} for a description of our metrics, the values are rounded. We evaluate on drifter observations from 2023 with magnitude $>0.25$ m/s. Best results are shown in bold font. }
\newcommand{\metricLines}[0]{{} & \multicolumn{2}{m{3.2cm}}{\textbf{Correct ang. \%}} & \multicolumn{2}{m{3.2cm}}{\textbf{Correct mag. \%}} & \multicolumn{2}{m{3.2cm}}{\textbf{MEVA (cm/s)}} \\  &  \textbf{T+1}  & \textbf{T+7} & \textbf{T+1} & \textbf{T+7} & \textbf{T+1} & \textbf{T+7} \\ \hline
}

\newif\ifDispComments
\DispCommentstrue

\ifDispComments
    \newcommand{\hannah}[1]{\textcolor{blue}{[Hannah: #1]}}
    \newcommand{\pierre}[1]{\textcolor{violet}{[Pierre: #1]}}
    \newcommand{\dominique}[1]{\textcolor{red}{[Dominique: #1]}}
    \newcommand{\anastase}[1]{\textcolor{orange}{[Anastase: #1]}}
    \newcommand{\theo}[1]{\textcolor{teal}{[Theo: #1]}}
    \newcommand{\ines}[1]{\textcolor{olive}{[Ines: #1]}}
    \newcommand{\amelie}[1]{\textcolor{cyan}{[Amelie: #1]}}
\else
    \newcommand{\hannah}[1]{}
    \newcommand{\pierre}[1]{}
    \newcommand{\dominique}[1]{}
    \newcommand{\anastase}[1]{}
    \newcommand{\theo}[1]{}
    \newcommand{\ines}[1]{}
    \newcommand{\amelie}[1]{}
\fi
\usepackage{siunitx}

\title{ORCAst: Operational High-Resolution Current Forecasts}

\author{%
  Pierre Garcia\thanks{These authors contributed equally to this work.} \\
  Amphitrite; Sorbonne Université, CNRS, LIP6 \\
  \And
  Inès Larroche\footnotemark[1] \\
  Amphitrite \\
  \And
  Amélie Pesnec\footnotemark[1] \\
  Amphitrite \\
  \And
  Hannah Bull\footnotemark[1] \\
  Amphitrite \\
  \And
  Théo Archambault \\
  Amphitrite \\
  \And
  Evangelos Moschos \\
  Amphitrite \\
  \And
  Alexandre Stegner \\
  Amphitrite \\
  \And
  Anastase Charantonis \\
  Inria, Sorbonne Université \\
  \And
  Dominique Béréziat \\
  Sorbonne Université, CNRS, LIP6 \\
}

\begin{document}

\maketitle

\begin{abstract}
We present ORCAst, a multi-stage, multi-arm network for Operational high-Resolution Current forecAsts over one week. Producing real-time nowcasts and forecasts of ocean surface currents is a challenging problem due to indirect or incomplete information from satellite remote sensing data. Entirely trained on real satellite data and in situ measurements from drifters, our model learns to forecast global ocean surface currents using various sources of ground truth observations in a multi-stage learning procedure. Our multi-arm encoder-decoder model architecture allows us to first predict sea surface height and geostrophic currents from larger quantities of nadir and SWOT altimetry data, before learning to predict ocean surface currents from much more sparse in situ measurements from drifters. Training our model on specific regions improves performance. Our model achieves stronger nowcast and forecast performance in predicting ocean surface currents than various state-of-the-art methods. 
\end{abstract}

\section{Introduction} \label{sec:intro}

Ocean currents are an important component of Earth's climate system, influencing weather patterns, marine ecosystems, and global heat distribution. Nevertheless, accurate reconstructions and forecasts of ocean surface current fields is a challenging task. There are two main types of observational data for estimating ocean surface currents: in situ measurements, for example from drifters, and satellite observations, for example from satellite altimetry and from multispectral optical satellites. These observational data can be assimilated in numerical models or used in data-driven methods to reconstruct and forecast ocean currents. 

In situ measurements, such as those from drifter buoys transported by the currents, allow direct measurements of ocean surface currents \citep{drifterscmems, driftersgdp}. However, drifters are sparse and unevenly distributed across the global ocean. Drifters are commonly used for evaluating the performance of models for estimating ocean currents~\citep{kugusheva_ocean_2024,martin_deep_2024,ODC}. In our work, we use drifters both as ground truth data to fine-tune our model, as well as to evaluate our method. Other sources of in situ data, for example from instruments on board ships, are generally not publically available, but limited examples of ship routes can be used to evaluate the performance of ocean current nowcasts and forecasts~\citep{kugusheva_ocean_2024}.

Satellite altimetry measures sea surface height (SSH), which is related to surface currents through the geostrophic balance approximation, i.e. an equilibrium between the Coriolis and pressure forces. Far from the Equator, geostrophic currents are the main component of the total ocean surface current. Since their first launch in 1978, nadir-pointing altimeters allow low resolution estimation of SSH through sparse along-track measurements~\citep{tapley1982seasat}. A widespread method of interpolating these sparse measures of SSH is DUACS, the Data Unification and Altimeter Combination System~\citep{taburet_duacs_2019}. This method is a global optimal interpolation, a linear interpolation method with optimally chosen coefficients of covariance matrices with respect to expert knowledge and historical data. However, due to the imited by the number of along-track measurements, the global maps of SSH provided by this optimal interpolation have a coarse resolution from 1/4° to 1/8° in specific regional areas. In February 2022, a new satellite called Surface Water and Ocean Topography (SWOT) was launched, providing high-resolution SSH data, below the kilometer, with a 120~km swath, significantly increasing the potential of satellite altimetry in measuring small-scale structures~\citep{morrow2018fine}.

In addition to SSH, ocean currents also have a physical signature on other remote sensing variables such as Sea Surface Temperature (SST) and chlorophyll-a (CHL). SST and CHL are passive tracers, or scalar quantities transported by the currents in via advection dynamics. They are measured by multispectral optical satellites, which allows for wide data coverage and high-resolution images (300m to 1km) when no clouds are present~\citep{emery_avhrr_1989}. Numerous studies demonstrate the importance of including multiple physical variables in reconstructing reliable surface currents~\citep{thiria_downscaling_2023,fablet_inversion_2024,martin_synthesizing_2023,archambault_learning_2024,kugusheva_ocean_2024,ciani_estimating_2024}.

\begin{figure}
    \centering
    \includegraphics[width=\linewidth]{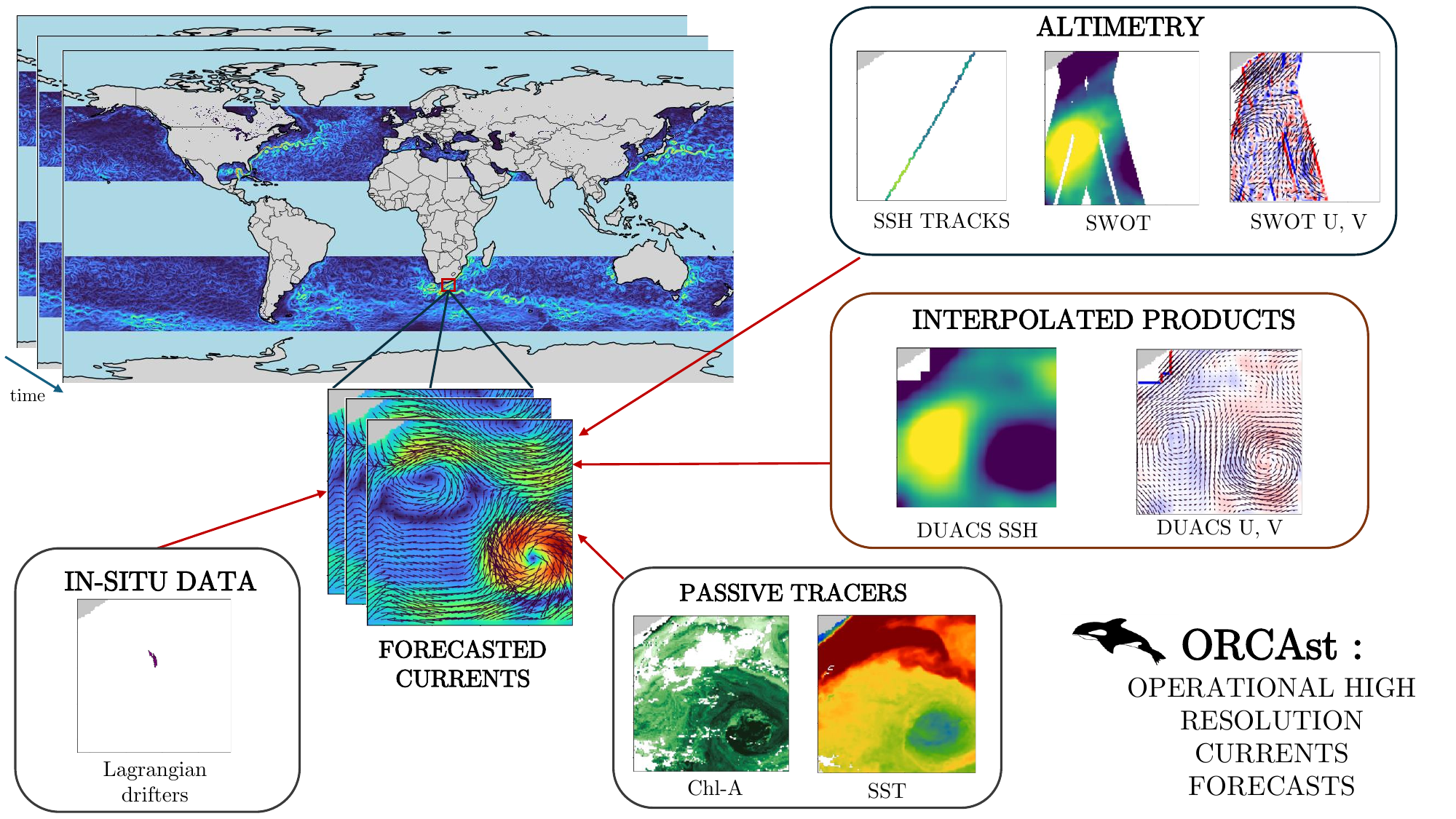}
    \caption{\hresvthree~is an operational neural network model that achieves state-of-the-art forecasts of ocean surface currents in extratropical latitudes. It is trained exclusively on observational data, including drifters, SWOT and nadir altimetry, SST and CHL observations, in a multi-stage training process. }
    \label{fig:teaser}
\end{figure}

There are two main methods of forecasting ocean surface currents using available drifter observations or observations from satellites: numerical models based on physical equations and data-driven approaches. Ocean Global Circulation Models (OGCMs) simulate ocean circulation using differential equations describing fluid dynamics~\citep{tonani_status_2015}. Starting from an estimation of the ocean state, OGCMs can forecast key physical variables. Initial conditions of the model can be estimated by assimilating observational data from satellites or in situ sensors. For example, the Mercator global forecasting system \citep{drevillon_godaemercator-ocean_2008,mercatordata} produces a global 10-day forecast of the ocean, by using the NEMO 3.6 model \citep{gurvan_nemo_2017} and by assimilating satellite observations of SSH, SST, Sea Ice Concentration, as well as in situ measurements. Physical simulations produce physically realistic fields at the cost of complex hyperparameter tuning and computationally expensive inference, although some methods can be used to reduce the computational burden of data assimilation \citep{le_guillou_regional_2023, ubelmann_simultaneous_2022}. 

Data-driven approaches, such as deep learning methods, use large quantities of data to learn the evolution of the physical system, rather than modelling the system dynamics through explicit equations. Deep neural networks have demonstrated high accuracy both in forecasting atmospheric evolution \citep{lam_learning_2023,chen_fuxi_2023}, and in forecasting ocean evolution \citep{wang_xihe_2024,aouni2024glonet}. However, the aforementioned approaches are trained on reanalysis data from numerical models, and not purely on observational data. More closely related to our work, various methods produce forecasts or time series of SSH and ocean surface currents using either numerical simulations or real data for training, or a mix of both. \citet{filoche2022} introduce a loss computed on nadir SSH observations only, and use it to perform a variational interpolation with a deep image prior. \citet{archambault2023visapp} extends this idea to leverage SST information. 
\citet{martin_synthesizing_2023} present a self-supervised neural network to perform delayed time SSH interpolation from partial SSH nadir-pointing and SST satellite observations. Using a related approach, \citet{archambault_pre-training_2024} shows that multi-stage training, i.e., using different datasets successively in the learning phase, increases SSH reconstruction quality. However, all these methods produce a delayed-time series of SSH fields, and must be adapted to forecast ocean surface currents.

In this work, we introduce \hresvthree, a data-driven method for Operational, high-Resolution Current forcAsts, illustrated in Figure~\ref{fig:teaser}. The novel features of this model are the following:
\begin{samepage}
\begin{itemize}
    \item \hresvthree\ is an operational neural network model to forecast SSH and ocean surface currents, without being limited to the geostrophic approximation, achieving state-of-the-art performance
    \item \hresvthree\ is trained purely on observational data, including drifters, SWOT and nadir altimetry, as well as SST and CHL observations,
    \item \hresvthree\ learns high-resolution ocean surface currents through a three-phases training procedure, which progressively refines forecasts with ground-truth data of increasing quality. 
\end{itemize}
\end{samepage}

The article is structured as follows: in Section~\ref{sec:data}, we describe the in situ and satellite data used in this study. Section~\ref{sec:method} provides a detailed explanation of the architecture of our model \hresvthree\ and our evaluation strategy. In Section~\ref{sec:results}, we present the experimental results demonstrating the operational benefits of \hresvthree. Finally, in Section~\ref{sec:discussion}, we discuss the conclusions and offer perspectives for future work.

\section{Data}\label{sec:data}

Ocean surface currents can be estimated from direct and indirect measurements from different sources of in situ and satellite remote sensing data. In this section, we provide details on the different data sources used to train or evaluate our model: in situ measurements from drifters and from instruments on board ships (Sec.~\ref{data:insitu}), SSH measurements from nadir-pointing altimeters and SWOT, as well as derived geostrophic currents (Sec.~\ref{data:ssh}), in addition to visible and infrared satellite images of chlorophyll and sea surface temperature (Sec.~\ref{data:sst}). We also describe our data preprocessing pipeline (Sec.~\ref{data:processing}). 

\subsection{In situ measurements}\label{data:insitu}
We use two types of in situ measurements: sparse observations from drifters and estimations of ocean currents from instruments on board ships. 

\begin{figure}
    \centering
    \includegraphics[trim=0cm 4cm 0cm 0cm, clip,  width=.99\linewidth]{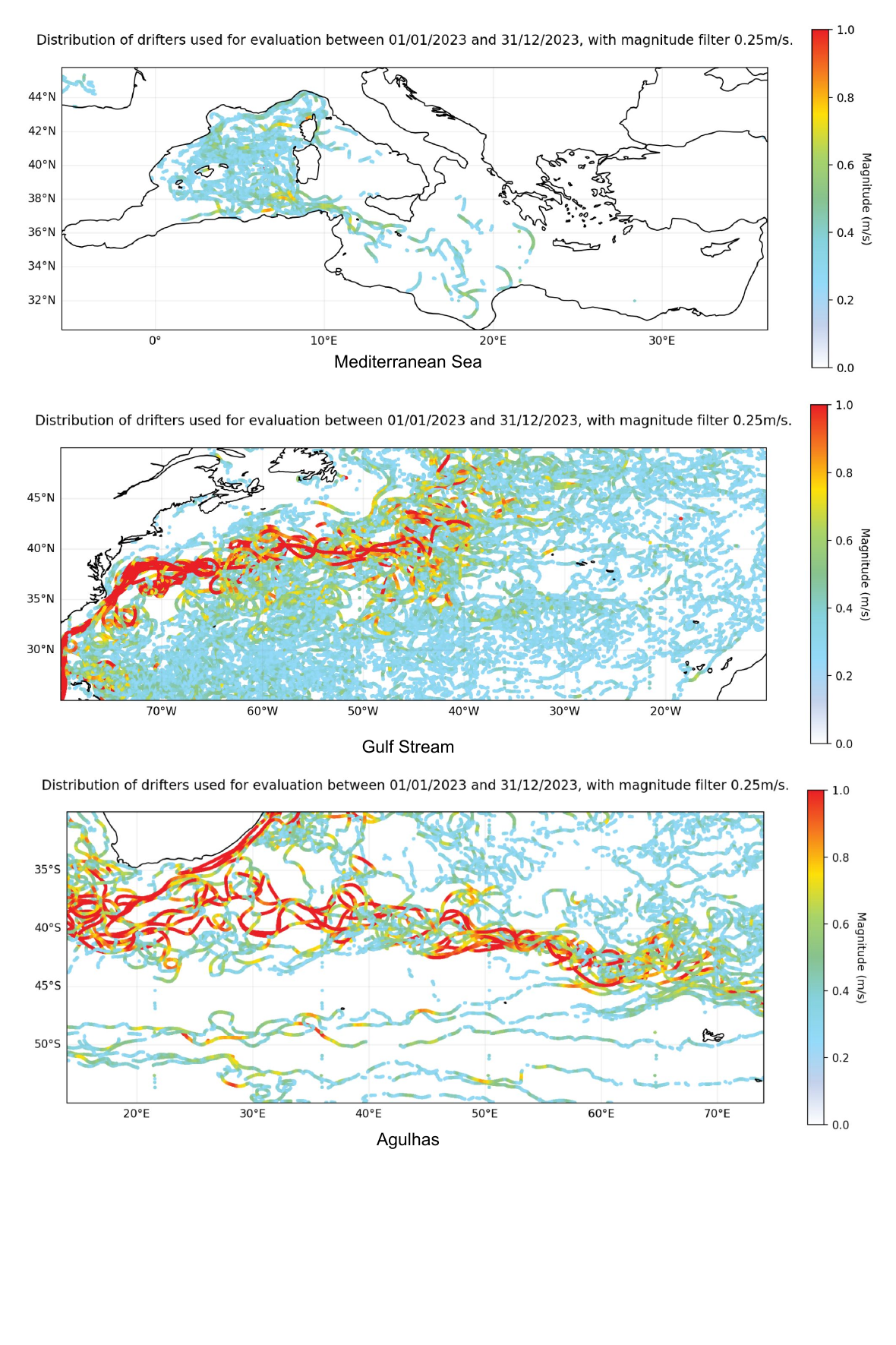}
    \caption{In situ observations of currents from drifters in three key regions in 2023, used for evaluation.}
    \label{fig:drifterRegionsVal}
\end{figure}

\textbf{Lagrangian drifters} are buoys equipped with a drogue and a GPS, and are advected by ocean currents. Our dataset contains hourly measurements of global drifters \citep{drifterscmems,driftersgdp,elipot2016globaldrifter}. In order to smooth noise, for instance due to surface winds and waves, we compute moving averages of observations over periods of 24 hours. Drifters are sparsely and non-uniformally distributed, which can bias learning and evaluation. Figure~\ref{fig:drifterRegionsVal} illustrates the spatial distribution of drifters in the Mediterranean, Gulf Stream and Aghulas region in 2023.

\textbf{On-board instruments from ships} also estimate the effect of currents, by measuring both Speed Through Water (STW) and Speed Over Ground (SOG). STW can be measured with instruments such as Wavex or Acoustic Doppler Current Profiler (ADCP), and SOG can easily be measured by regular GPS localisation. The difference between these measurements, $\mathrm{SOG} - \mathrm{STW}$, is an indicator of the effect of currents felt by ships along their trajectory. This data is limited to a specific examples of ship voyages and thus only used for evaluation purposes.

\subsection{Satellite altimeter measurements}\label{data:ssh}
Far from the Equator, the ocean circulation is primarily driven by its geostrophic component, which can be derived from SSH. Two main types of satellite measure the SSH: nadir-pointing altimeters and the Ka-band Radar Interferometer (KaRIn) on board the SWOT satellite. 
Numerous interpolation methods exist to produce complete 
estimations of SSH and geostrophic currents such as \citet{taburet_duacs_2019} DUACS and \citet{scott_neurost} NEUROST. 

\textbf{Nadir-pointing altimeters} such as those used in various satellite missions like Sentinel-6A, Jason-3, and Cryosat-2, measure SSH along the satellite's track with a spatial resolution of approximately 7~\unit{\kilo\meter}. The data is processed through the DUACS protocol, which ensures consistency across multiple altimeter missions by standardizing them to a reference mission~\citep{sshtracks}. 

\begin{figure}[ht]
    \centering
    \begin{minipage}{.47\textwidth}
        \centering
    \includegraphics[width=\linewidth]{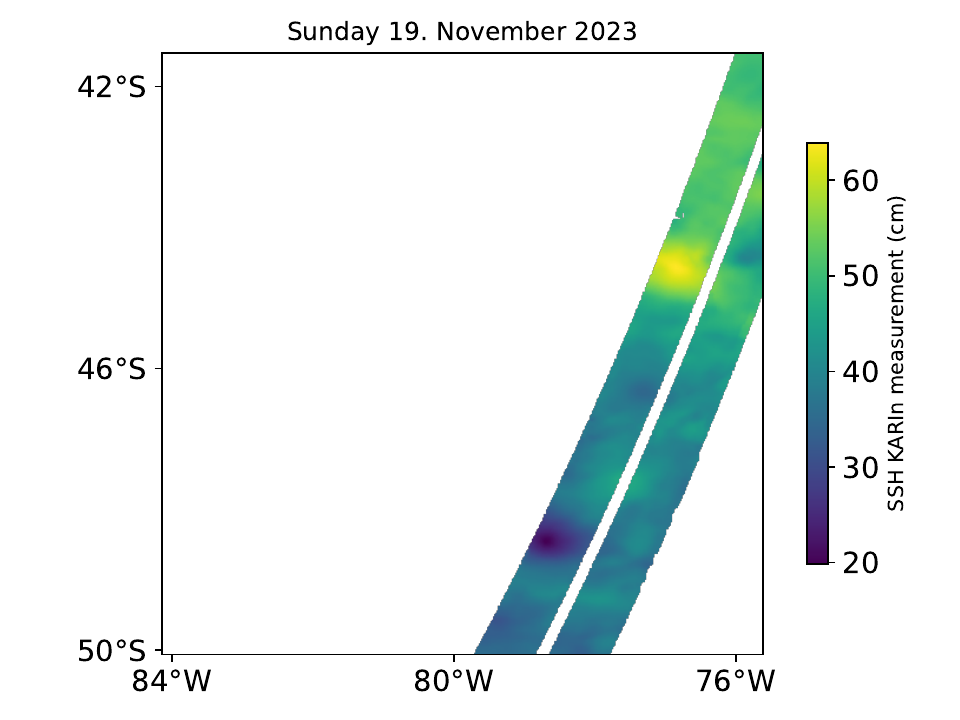}
    \caption{SWOT SSH observation, with visible cyclonic and anti-cyclonic eddies at high resolution. SWOT consists of two 50~km wide bands, separated by a 20~km gap covered by traditional nadir altimetry instruments. }
        \label{fig:figure1}
    \end{minipage}\hfill
    \begin{minipage}{.47\textwidth}
        \centering
        \includegraphics[width=\linewidth]{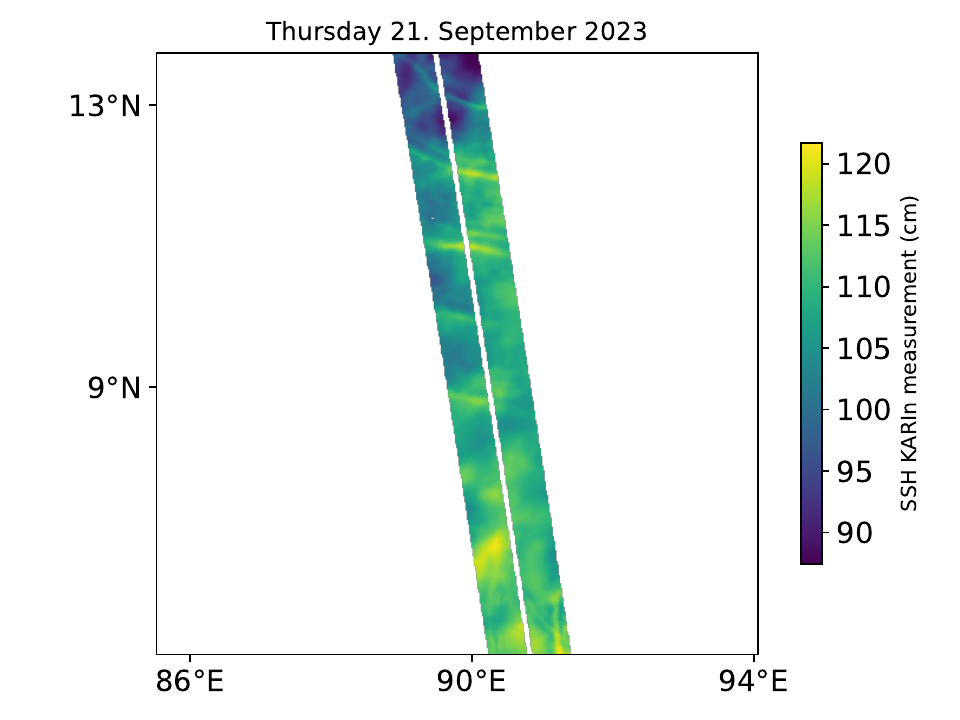}
        \caption{SWOT SSH observation, with visible internal gravitational waves. In particular, close to the Equator, it is more difficult to estimate ocean surface currents from SSH as a result of the reduced Coriolis effect. }
        \label{fig:figure2}
    \end{minipage}
\end{figure}

\textbf{KaRIn altimetry from SWOT} provides exciting opportunities for precise reconstruction of the SSH. 
SWOT provides 120~\unit{\kilo\meter}-wide swaths of high-resolution observations~\citep{DATAswot}.
This precision allows observation of mesoscale/submesoscale SSH patterns. However, only one satellite is presently in orbit, its revisit time period is 21 days, which leaves important areas unobserved. We have access to around one year of SWOT observations, compared to decades of nadir-pointing altimeter data. Figures~\ref{fig:figure1} and~\ref{fig:figure2} show examples of fine-scale eddies visible from SWOT, but also internal gravitational waves close to the Equator. This illustrates SWOTs potential in estimating ocean surface currents, but also its limitations, particularly in equatorial latitudes. 

\textbf{DUACS} is a Level 4 (L4) product, providing a low-resolution ($0.125$°) mapping of SSH through optimal interpolation~\citep{taburet_duacs_2019} of nadir altimetry. The geostrophic currents can then be derived from this mapping, and are available at~\citep{medduacs,globalduacs}. For evaluation and stage 1 training, we use a delayed-time DUACS product, which uses 6 days of future observations from nadir tracks. 

\textbf{NeurOST SSH} is a state-of-the-art method for predicting global SSH at $0.1$° resolution. \citet{martin_deep_2024} train a neural network to predict nadir-pointing along-track SSH from SST and SSH inputs, using only observations in a self-supervised manner. 
This product is produced in delayed-time, integrating 14 days of future observations\footnote{The authors of NeurOST have informed us that there were some issues in production of this data, potentially reducing quality, and that there will be a new release in the near future. We were unaware of this issue at the time of analysis, and have used the NeurOST data as provided in November 2024.}.

\subsection{Satellite imagery}\label{data:sst}
SST and CHL concentration are passive tracers of ocean currents, as water motion advects these scalar variables. Small-scale structures in ocean surface currents are visible using high-resolution satellite imagery of SST and CHL. However, both infrared and visible waves are blocked by clouds, and so information provided from these sensors is partial. Although methods exist for interpolating SST and CHL, we use L3 and L2 products and allow our model to learn to combine these different sources of information from past days. 

\textbf{SST} is measured using infrared sensors at a high spatial resolution. We use the daily L3 products~\citep{globalsst,medsst}, with a resolution of $0.02$°. 

\textbf{Chlorophyll-a} (CHL) is measured using data from the Ocean and Land Colour Instrument (OLCI) aboard Sentinel-3 satellites, which provides ocean color geophysical products. The chlorophyll product is provided using the OC4Me algorithm~\citep{oreilly_ocean_1998}. This algorithm calculates the concentration of algal pigments by analyzing water-leaving reflectances across 21 spectral bands. The chlorophyll concentrations are provided in log10-scaled values. We use data available at 1.2 km resolution~\citep{chldata}. 

\subsection{Data processing pipeline}\label{data:processing}
Our data processing pipeline interpolates all data sources into a common grid at resolution 1/30°. We compute means and standard deviations of the key variables - SSH, U and V components of ocean currents, SST and CHL - at a spatial resolution of 2° and a temporal resolution of 1 week. To estimate the means and standard deviations of SSH and ocean surface currents, we use the DUACS L4 product. We normalise all variables using these values to remove average seasonal and spatial effects.

\section{Method}\label{sec:method}

\hresvthree~is a deep learning model to forecast ocean currents. Our model inputs $T$ timesteps of past and present satellite observations of SSH and optical satellite data, and outputs $\tau$ future timesteps of SSH and the U and V components of ocean surface currents. We learn the parameters $\theta$ of the neural network $f
_\theta$, shown in Equation~\ref{eq: forecast},
\begin{equation}
    f_\theta\left(X^{0 : T}\right)=\widehat{Y}_{\mathrm{SSH}}^{T+1 : T+\tau},\widehat{Y}_{\mathrm{U, V}}^{T+1 : T+\tau} \label{eq: forecast}
\end{equation}
where $X$ designates past and present observations and $\widehat{Y}$ the forecasted fields. The inputs and outputs are gridded regional crops of dimension $h\times w$ and at identical spatial resolutions. In practice, we input $T=11$ days of past and present satellite observations and output $\tau=7$ future days of satellite observations. Inputs $X$ and outputs $\widehat{Y}$ are daily images of size 128$\times$128 at 1/30° resolution. 

In Section~\ref{method:architecture}, we present our multi-arm encoder-decoder model architecture and in Section~\ref{method:multstagetraining}, we describe our three-stage training procedure. Section~\ref{method:geozones} provides the geographic zones used for training and evaluation, and Section~\ref{method:metrics} outlines our evaluation metrics and our baseline comparisons. 

\subsection{Model architecture}\label{method:architecture}

\begin{figure}
    \centering
    \includegraphics[width=1\linewidth]{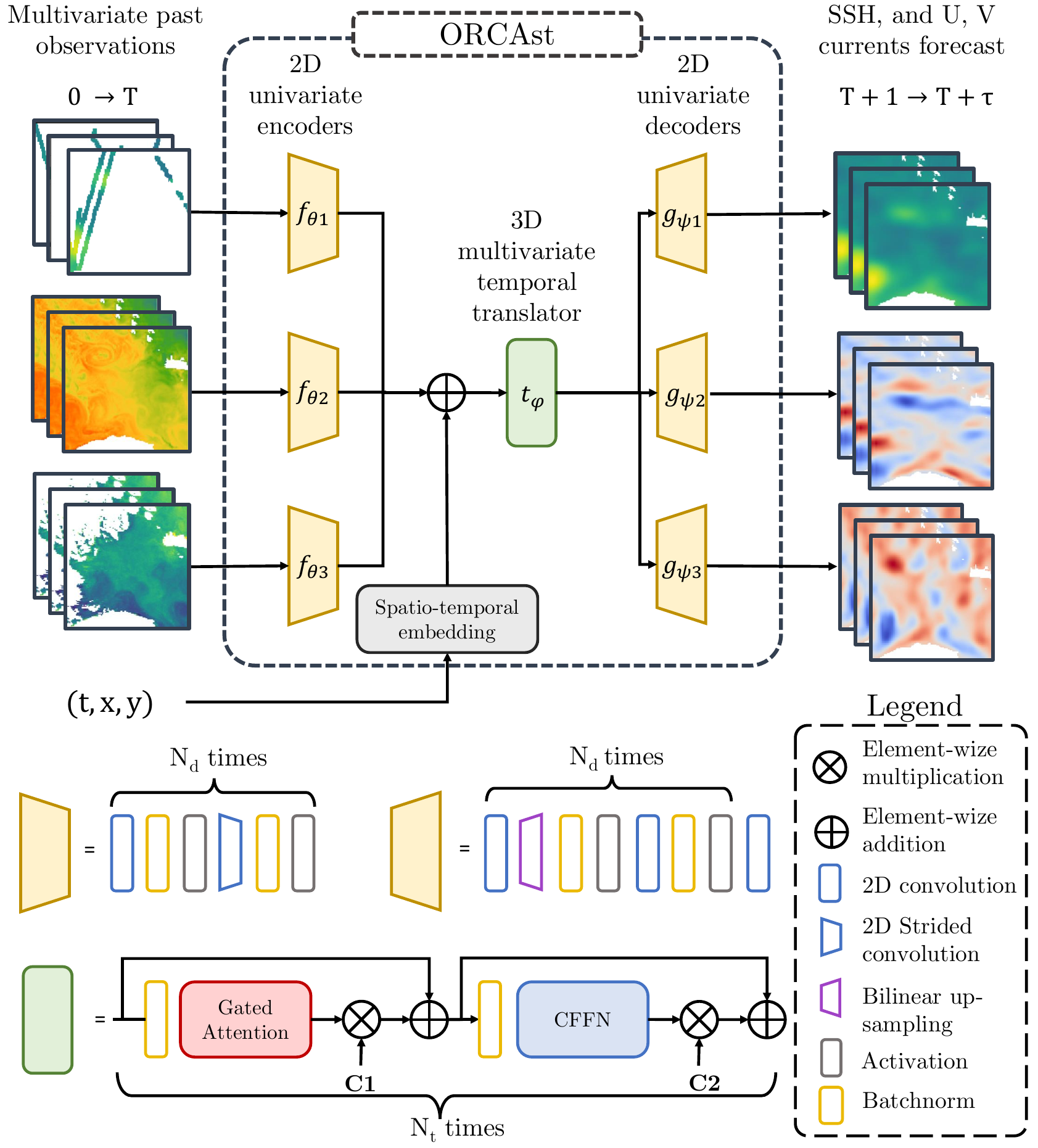}
    \caption{Overview of~\hresvthree. Our models inputs temporal sequences of multivariate satellite observations (here SSH, SST, CHL) from $t=0$ to $t=T$. Each variable is inputted to a different spatial encoder, which extracts information from each timestep separately. A spatio-temporal positional embedding is applied to inform the network of geographical coordinates and seasonal information. A spatio-temporal translator $t_\varphi$ is used to produce the forecast encodings. Finally, each timestep is decoded independently with three univariate decoders $g_\psi$, forecasting SSH, $U$ and $V$ components separately.}
    \label{fig:compass_model}
\end{figure}

Our model architecture, illustrated in Figure~\ref{fig:compass_model}, is based on SimVP, a flexible encoder-decoder architecture initially designed for video prediction~\citep{gao_simvp_2022}. A similar architecture is used to reconstruct a time series of SSH from nadir altimetry and SST inputs by~\citet{martin_deep_2024}. The \hresvthree\ model has the following key features:
\begin{itemize}
    \item one 2D encoder per input variable, allowing the model to best exploit each variable and timestep with fewer parameters;
    \item one 2D decoder per output variable, allowing us to remove decoder arms during training (e.g. in Stage 3 of our training procedure, described in Section~\ref{method:multstagetraining}), and also reducing the number of parameters;
    \item we replace the deep inception module from \citet{gao_simvp_2022} with a Gated Spatio Temporal attention (GSTa) based module for better efficiency;
    \item we add a spatio-temporal positional encoding module before the temporal translator, which allows the model to learn regional and seasonal phenomena.
\end{itemize}

\textbf{2D univariate encoders.}
The network encodes each variable and timestep separately, using different encoders ($f_{\theta_i}$ in Figure~\ref{fig:compass_model}) and dividing the spatial dimensions by four. The same encoder is applied to each timestep, but different encoders encode each physical variables, enabling the network to learn the different characteristics of each input data source. The encoders first apply a $4\times 4$ convolution, followed by a GroupNorm from \citet{wu_group_2020} and Gaussian Error Linear Unit (GELU) activation function. Then a strided convolution divides spatial dimensions by 2. This module is then repeated to produce the final latent space. 

\textbf{Spatial-temporal embedding.}
After the encoder, we apply a positional encoding, i.e. a learnable distribution shift of the latent image informing the network about the spatio-temporal coordinates of the patch (see Appendix~\ref{apx:pos_enc}). The temporal dimension is then placed in the variable dimension, meaning that the model can subsequently access temporal dependencies through the variable dimension.

\textbf{3D multivariate temporal translator.}
The Gated Spatio Temporal Attention (GSTa) module ($t_\varphi$ in~\ref{fig:compass_model}) is composed of a sequence of gated attention sub-blocks. As the the temporal dimension is merged with variables, GSTa is able to learn the dynamics of the timeseries by multi-variables convolutions. A block of the GSTa model consists of a gated attention module, followed by a Convolutional Feed Forward Network (CFFN) from \citet{wang_pvt_2022}, used with residual connections and learned rescaling, schematised by $t_\varphi$ in Figure~\ref{fig:compass_model}. This module contains dilated convolutions and an attention gating processes, allowing the model to learn long range spatial interactions. For further details about the GSTa module, we refer the reader to~\citet{metaFormer}.

\textbf{2D univariate decoder.}
After the multivariate temporal translator, we apply three 2D univariate decoders in order to decode SSH, the meridional current component $U$, and the zonal current component $V$. Similarly to the encoders, the decoders $g_{\varphi_i}$ apply the same operation to each timestep, but have different weights for each variable. They are composed of $4\times 4$ convolutions followed by bilinear upsampling layers. These operations are repeated twice to obtain the original image dimensions.

\subsection{Multi-stage training}\label{method:multstagetraining}

Training a model that leverages the diverse range of oceanographic observations — such as drifters, satellite altimetry, and satellite imagery — is challenging due to the differing temporal coverage and resolutions of these data sources. One might consider developing a model to forecast ocean currents and sea surface height (SSH) by directly comparing predictions to all available measurements, using drifter velocities as targets for currents and combining observations from nadir and SWOT altimetry for SSH forecasting. However, this approach presents several significant challenges. First, observations from different sources are not always consistent; for example, inconsistencies have been identified in some SWOT and nadir crossover points, as discussed in Appendix~\ref{apx:karin_nadir_bias}. Second, the temporal coverage varies greatly between data sets: SWOT provides only about a year of data, while nadir altimetry has accumulated observations over several decades~\citep{taburet_duacs_2019}. Third, the density of observations differs across sources, with drifter data being far sparser than satellite measurements, making it difficult to appropriately weight each data source during model training.

\begin{figure}
    \centering
    \includegraphics[width=\textwidth]{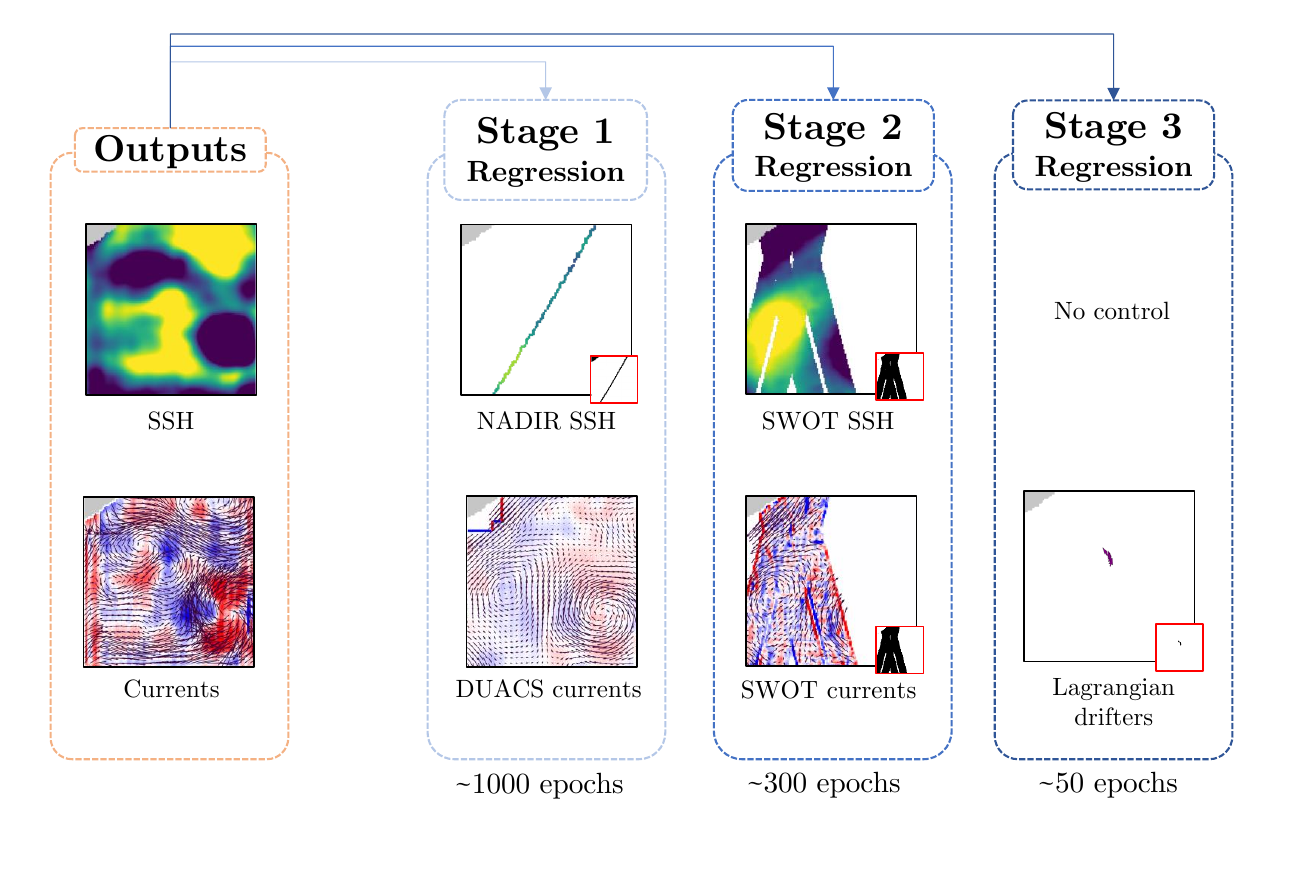}
    \caption{ORCAst training stages. From left to right, we first show an output example of SSH and $U$ and $V$ components of ocean surface currents, represented as a vector field with colours denoting vorticity. Stage 1 shows nadir and DUACS regression targets, where the small red box illustrates the mask of pixels taken into account when computing the loss. Stage 2 shows SWOT regression targets, with associated masks on the loss function in the small red box. Stage 3 shows drifter regression target, with masks on the loss function in the small red box. Note that in Stage 3 we freeze the SSH decoder and train the model to learn only $U$ and $V$. }
    \label{fig:multi-stage}
\end{figure}

Considering these difficulties, we train our model using a 3-stage training strategy, illustrated in Figure~\ref{fig:multi-stage}. The goal of this strategy is to learn a first approximation of SSH and ocean currents using nadir-pointing SSH and derived geostrophic currents from interpolated SSH fields as ground truth, and to progressively refine the learning using scarcer, but more precise or direct observations (SWOT and drifters). The inputs of the neural network are the same at each stage: nadir SSH and SST. We also include experiments on using additional CHL or SWOT observation as inputs. Our strategy contains 3 stages of training:

    \begin{enumerate}
    \item Our model estimates along-track nadir SSH, and DUACS delayed-time geostrophic currents. As the nadir pointing observations are sparse, we first perform a masking operation before computing the loss, so that the network can be trained with incomplete fields. For further details about this methodology we refer the reader to~\citet{filoche2022,archambault2023visapp, martin_synthesizing_2023,archambault_learning_2024}. This corresponds to Stage 1 of Figure~\ref{fig:multi-stage}.
    \item Our model estimates high-resolution SWOT data of SSH and associated geostrophic currents. As SWOT produces large bands of observations, it is possible to compute spatial gradients and thus retrieve the geostrophy directly from observations without interpolation methods. However, SWOT provides partial spatial coverage, so we apply the masking operation to both SSH and surface current outputs of \hresvthree\ before computing the loss. This corresponds to Stage 2 of Figure~\ref{fig:multi-stage}.
    \item The final stage involves finetuning only the output currents, whilst freezing the SSH decoder. In this stage, the regression of \hresvthree\ currents fits to sparse but direct drifter measurements of $U$ and $V$, thereby allowing the model to learn from the total currents rather than a geostrophical approximation. This corresponds to Stage 3 of Figure~\ref{fig:multi-stage}.
\end{enumerate}

At each stage, we train the model using MSE loss. In order for the model to learn to reconstruct structures and fast-moving ocean currents, we weight the MSE loss function proportionally to the magnitude of ocean currents in the target observations, following~\citet{kugusheva_ocean_2024}. We use an Adam optimizer~\citep{kingma2014adam} with the learning rate set to $10^{-3}$ and weight decay to $10^{-3}$ in Stage 1, and resume training by reducing the learning rate by a factor of 10 in Stages 2 and 3. We set each epoch to 1000 randomly chosen patches. During the three stages, the model is trained using 1000, 200 and 50 epochs respectively. When training in the smaller area of the Mediterranean Sea (described in Section~\ref{method:geozones}), we train for 500, 200 and 25 epochs respectively, over the 3 stages. The number of epochs was chosen via a validation set. We merge patches using Gaussian kernel-weighted averaging, as detailed in Appendix A of \citet{callaham_robust_2019}.

Stages 1 and 2 of our analysis are trained using geostrophic currents, which are the dominant components of ocean surface currents far from the Equator. Tropical latitudes, where the Coriolis effect is weaker and other fast ageostrophic dynamics such as internal gravitation waves (Figure~\ref{fig:figure2}) or equatorial Rossby and Kelvin waves become more significant, are excluded from our analysis. See Figure~\ref{fig:emb_clust} for a further illustration of the importance of latitude. 

\subsection{Geographic zones and time periods}\label{method:geozones}

The study area is divided into several subregions, defined in Table~\ref{tab:region_definitions}. We exclude equatorial regions as we train using geostrophic currents in Stages 1 and 2, which are only a reasonable approximation of the total currents far from the Equator. In order to compare the performance of regional-specific models and global models, we train our model on three regions: the Mediterranean Sea, the Gulf Stream and the Agulhas region (Table~\ref{tab:region_definitions}). 

\begin{table}[h!]
    \centering
    \caption{Geographical definitions of oceanic regions by latitude and longitude.}
    \begin{tabular}{ccc}
        \hline
        \textbf{Region} & \textbf{Latitude Range} & \textbf{Longitude Range} \\ \hline
        {Global (minus Equator and Poles)} & -60° to -20° and 20° to 60° & -180° to 180° \\ \hline
        {Mediterranean Sea} & 30° to 46° & -6° to 36° \\ \hline
        {Gulf Stream} & 20° to 45° & -99° to -34° \\ \hline
        {Agulhas} & -55° to -30° & 14° to 74° \\ \hline
    \end{tabular}
    \label{tab:region_definitions}
\end{table}

We train our model using data from 2016 to mid-December 2018 and from mid-January 2020-2022 during Stages 1 and 3. In Stage 2, we use 6 months of available SWOT data from 2024, until July. We use 2023 to evaluate our model's performance. 

\subsection{Evaluation metrics and baselines}\label{method:metrics}

\begin{figure}
    \centering
    \includegraphics[width=\textwidth]{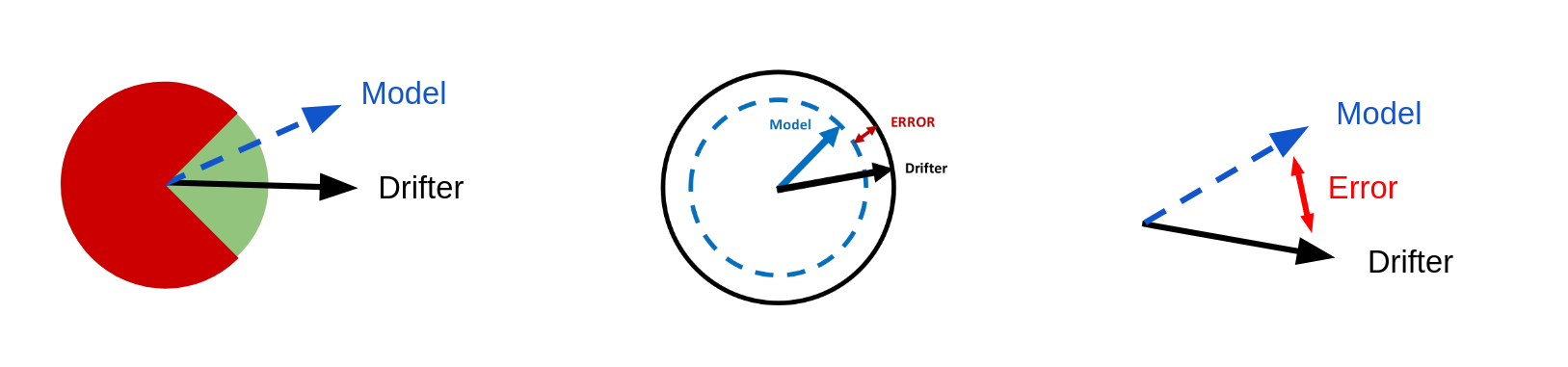}
    \caption{Simplified illustration of our three metrics, from left to right: angle error, magnitude error and error vector amplitude when compared to drifter measurements.}
    \label{fig:metrics}
\end{figure}

We evaluate our results on sparse in situ observations collected from Lagrangian drifters in the year 2023 (see Figure~\ref{fig:drifterRegionsVal}). As we are interested in detecting eddies and other structures with fast-moving currents, we restrict our drifter evaluation set to observations above 0.25~m/s. In the following, let $\hat{w}$ be the velocity vector predicted by the model at the position of the drifter, and $\textit{w}_{drifter}$ be the velocity vector observed by the drifter. We measure prediction quality using 3 metrics, illustrated in Figure~\ref{fig:metrics}:

\begin{samepage}
\begin{itemize}
\item{\textbf{Percentage of correct angles}} Let $\theta$ be the angle difference between vectors $\hat{w}$ and $\textit{w}_{drifter}$. Then:

$$ \theta= \frac{180}{\pi} \cos^{-1} \left(\frac{\hat{w}\cdot w_\text{drifter}}{\|\hat{w}\|\|w_\text{drifter}\|}\right) \in [0; 180]. 
$$

If $\theta \leq 45$°, we consider the current \textbf{direction} as \textbf{correct}. We then compute the percentage of correctly predicted angles in our drifter evaluation set. 
 
\item{\textbf{Percentage of correct magnitude}} \\ Let $\Delta M$ be the magnitude difference between $\hat{w}$ and $\textit{w}_{drifter}$. Then:

$$ 
\Delta M = | \|\hat{w}\| - \|w_\text{drifter}\| | \in \mathbb{R}^+.
$$
If $\Delta M \leq 2.5 $ \unit{\centi\meter/\second}, we consider the current \textbf{magnitude} as \textbf{correct}. We then compute the percentage of correctly predicted magnitudes in our drifter evaluation set. 

\item{\textbf{Mean Error Vector Amplitude (MEVA)}} \\ 
The Error Vector Amplitude is the velocity difference $\Delta V$ between $\hat{w}$ and $\textit{w}_{drifter}$, i.e.
$$
\Delta V = \|\hat{w}-w_\text{drifter}\| \in \mathbb{R}^+.
$$
We compute the mean of the Error Vector Amplitude, or the MEVA, across drifter observation in our evaluation set. 
\end{itemize}
\end{samepage}

These three metrics for our predicted ocean surface current forecasts are calculated at time 
$T+1$ and $T+7$ days. Our model performance is compared to three alternative methods: DUACS, NeurOST and Mercator.
As DUACS and NeurOST do not provide a forecasting method, we compare with persistence forecasting, i.e. compare our model's forecasts at time $T+7$ with DUACS and NeurOST predictions at time $T+1$. 
Note that our method is operational, we use only data up until time $T$ to predict forecasts of $T+1$ until $T+7$. We are thus at a disadvantage when comparing to 6-day delayed-time ocean surface currents from DUACS or 14-day delayed-time ocean surface currents from NeurOST. Mercator~\citep{drevillon_godaemercator-ocean_2008,mercatordata} provides operational near real-time forecasts of ocean surface currents, and we compare our ocean surface current forecasts to Mercator forecasts.

\section{Results}\label{sec:results}

In this section, we provide quantitative and qualitative results of our ocean current forecasts model, using independent in situ observations from drifters from 2023. Section~\ref{results:global} presents our results on the extratropical Global area and the importance of our proposed three-stage training procedure. Section~\ref{results:regional} details model performance when trained at a global scale or trained specifically on regional areas. In Section~\ref{results:inputdata}, we conduct ablation studies experimenting with CHL and SST inputs, as well as SWOT data as inputs. Section~\ref{results:quali} illustrates the qualitative performance of our model, compared to baselines. Finally, Section~\ref{results:ship} evaluates our model on real ship data, demonstrating a potential application of forecasting ocean currents in the maritime industry. 

\subsection{Global forecast}\label{results:global}

Table~\ref{tab:global_comp} presents the performance of our model, trained and evaluated over the extratropical Global area, using the metrics and baselines outlined in Section~\ref{method:metrics}. Our method consistently outperforms the baselines in both next-day ($T+1$) and 7-day ($T+7$) forecasts. For example, at $T+1$, our model achieves 85\% of correct angles, compared to 78\% for DUACS, 83\% for NeurOST, and 70\% for Mercator. Similarly, for $T+7$ forecasts, our model reaches 70\% of correct angles, outperforming DUACS (62\%), NeurOST (65\%), and Mercator (56\%). These results highlight that despite being an operational, near-real-time forecasting model, \hresvthree\ surpasses the delayed-time methods DUACS (6-day delay) and NeurOST (14-day delay) while also significantly improving over the near-real-time numerical model Mercator. This demonstrates the ability of our deep learning-based approach to effectively capture physical relationships from observational data.

\begin{table}
\caption{\textit{Evaluation on extratropical Global area.} On all metrics, our near-real-time model outperforms existing methods for estimating ocean surface currents at time $T+1$, including both delayed-time methods such as DUACS (6 days delay) and NeurOST (14 days delay) as well as near real-time numerical forecast methods such as Mercator. Moreover, our forecasts at time $T+7$ improve performance over Mercator's numerical forecast method and persistence forecasts of DUACS and NeurOST. \metricdesc\ }
\label{tab:global_comp}
\resizebox{\textwidth}{!}{\begin{tabular}{b{3.7cm}b{0.8cm}b{1.5cm}b{0.8cm}b{1.5cm}b{0.8cm}b{1.5cm}}
    \topline
    \metricLines
    DUACS (DT pers.)
    & 78 & 62 & 69 & 68 & 25 & 37 \\ \midline
    NeurOST (DT pers.)
    & 83 & 65 & 72 & 67 & 25 & 31 \\ \midline
    Mercator
    & 70 & 56 & 68 & 68 & 25 & 37 \\ \midline
    \hresvthree
    & \textbf{85} & \textbf{70} & \textbf{77} & \textbf{69} & \textbf{24} & \textbf{30} \\ \botline
\end{tabular}}
\end{table}

The contribution of the three training stages to the performance of \hresvthree\ is shown in Table~\ref{tab:stage_importance}, with each stage progressively enhancing the model's accuracy. During Stage 2, incorporating SWOT data allows the model to leverage high-resolution SSH observations, improving the percentage of correct angle predictions from 79\% in Stage 1 to 83\% in Stage 2. Fine-tuning in Stage 3 using in situ drifter measurements further refines the model's ability to capture ocean surface currents beyond the geostrophic component. Stage 3 boosts the percentage of correct angles to 85\%. Similarly, the 7-day ($T+7$) forecast accuracy improves steadily, with the percentage of correct angles increasing from 64\% in Stage 1 to 68\% in Stage 2, and reaching 70\% in Stage 3. The model also experiences a substantial improvement in magnitude performance across the stages. For next-day ($T+1$) forecasts, the percentage of correct magnitudes increases significantly from 59\% in Stage 1 to 73\% in Stage 2, and finally to 77\% in Stage 3. These results emphasize the critical role of each stage in enabling the model to effectively utilize diverse observational data for more accurate ocean current predictions.

\begin{table}[htbp]
\caption{\textit{Importance of 3-stage training.} Comparison of ORCAst performances on the extratropical Global area at each training stage, using nadir-pointing altimetry and SST as inputs. Each training stage brings additional improvement in the considered metrics. \metricdesc\ } 
\label{tab:stage_importance}
\resizebox{\textwidth}{!}{\begin{tabular}{b{3.7cm}b{0.8cm}b{1.5cm}b{0.8cm}b{1.5cm}b{0.8cm}b{1.5cm}}
    \topline
    \metricLines
    \hresvthree~\phaseOne
    & 79 & 64 & 59 & 57 & 30 & 50 \\ \midline
    \hresvthree~\phaseTwo
    & 83 & 68 & 73 & 68 & 27 & 34  \\ \midline
    \hresvthree~\phaseThree
    & \textbf{85} & \textbf{70} & \textbf{77} & \textbf{69} & \textbf{24} & \textbf{30} \\ \botline
\end{tabular}}
\end{table}

\begin{figure}
    \centering
    \includegraphics[width=0.9\linewidth]{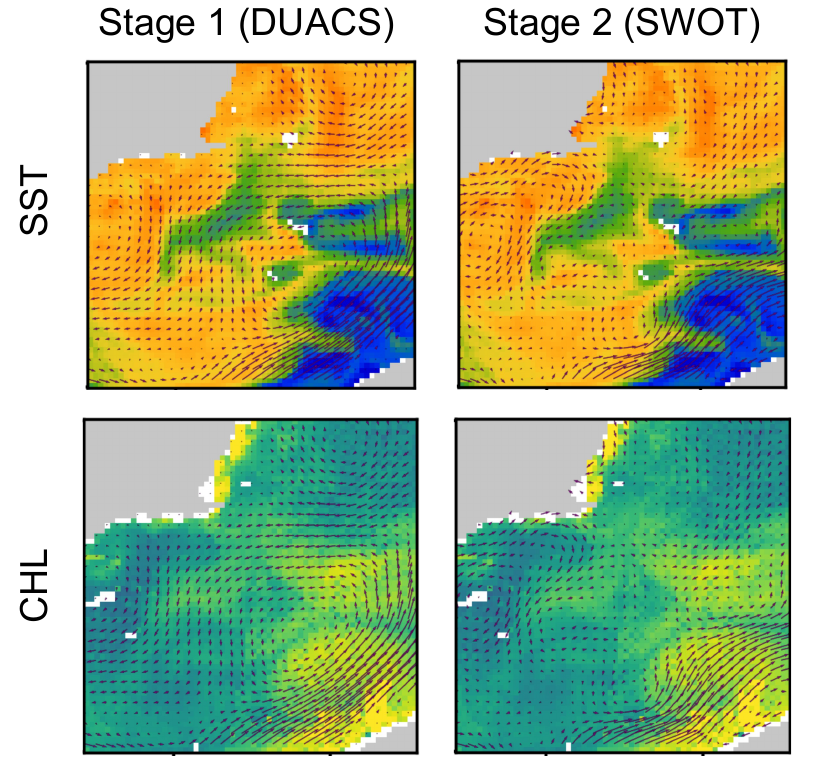}
    \caption{Example of performance of \hresvthree\ in the Mediterranean Sea on August 31, 2023, within latitude and longitude bounds of 35.0°N–39.3°N and 2.6°W–1.7°E. SST (top line) and CHL (bottom lines) are shown in the background. Black arrows denoted predicted ocean surface currents after Stage 1 (left column) and Stage 2 (right column) of the regional model trained in the Mediterranean Sea (see Table~\ref{tab:med}). In Stage 1, the model is trained to predict large-scale structures using nadir altimetry data. In Stage 2, the model is trained with fine-scale SWOT altimetry data, which allows it to be better rewarded for accurately predicting small-scale structures seen in the SST signature. }
    \label{fig:quali_twostages}
\end{figure}

The importance of Stage 2 in learning fine-scale structures is shown in Figure~\ref{fig:quali_twostages}. During Stage 1, the model is trained to predict large-scale ocean structures using nadir altimetry data, and is not penalised for failing to detect structures below the effective resolution of nadir altimetry ($\sim100$~km). In Stage 2, the model is further refined to predict smaller-scale structures, visible in the SST and CHL signatures, by training on high-resolution SWOT altimetry data with a lower effective resolution ($\sim15$~km). We refer to Appendix~\ref{apx:neurostAblation} for an ablation study on using NeurOST instead of DUACS in Stage 1. 

\subsection{Regional forecasts}\label{results:regional}

This section explores the differences between training models on a large, global scale and using region-specific models. Results in Tables~\ref{tab:med}, \ref{tab:gs}, and \ref{tab:agh} show that regional models consistently outperform the global model in the Mediterranean Sea, Gulf Stream region, and Agulhas region. Regional \hresvthree\ models benefit from focusing on the unique physical characteristics of their specific areas, while the global \hresvthree\ model must generalize across a wide variety of ocean dynamics. Although our global model incorporates geographical information through positional encoding, this alone is insufficient to fully resolve the nuanced differences across regions at the same capacity, as the model is still required to generalise across a much wider range of physical and environmental conditions, increasing the difficulty of the learning task.

For instance, the Mediterranean Sea is an enclosed basin with lower energy levels and distinct physical characteristics compared to the open ocean. The Mediterranean Sea also features smaller-scale structures and a reduced Rossby radius, which partly explains why the regional Mediterranean model achieves a significant improvement over the globally-trained model. As shown in Table~\ref{tab:med}, the regional model achieves 85\% correct angle predictions at time $T+1$, compared to 77\% for the global model, and 88\% of correct magnitudes at time $T+1$, compared to 78\% for the global model. For 7-day lead times, improvements are even more significant. At $T+7$, the Mediterranean model achieves 70\% correct angles, compared to 58\% for the global model, and 83\% correct magnitudes, compared to 59\% for the global model. This substantial improvement in both angle and magnitude accuracy underscores the advantage of focusing on the region's unique dynamics.

\begin{table}[htbp]
\caption{\textit{Evaluation in Mediterranean Sea.} Our model trained specifically on the Mediterranean Sea outperforms \hresvthree\ trained on the extratropical Global area, and also achieves higher accuracy than alternative methods. \hresvthree\ trained on the Global area outperforms DUACS, NeurOST and Mercator at time $T+1$, but it has slightly lower performance than persistence NeurOST for 7-day forecasts, although we note that NeurOST is a 14-day delayed-time method. \metricdesc\ } 
\label{tab:med}
\resizebox{\textwidth}{!}{\begin{tabular}{b{3.7cm}b{0.8cm}b{1.5cm}b{0.8cm}b{1.5cm}b{0.8cm}b{1.5cm}}
    \topline
    \metricLines
    DUACS (DT pers.)
    & 58 & 55 & 51 & 51 & 29 & 31 \\ \midline
    NeurOST (DT pers.)
    & 74 & 63 & 65 & 62 & 25 & 28  \\ \midline
    Mercator & 51 & 43 & 64 & 60 & 32 & 35 \\ \midline
    \hresvthree~Global
    & 77 & 58 & 78 & 59 & 26 & 30  \\ \midline
    \hresvthree~Regional 
    & \textbf{85} & \textbf{70} & \textbf{88} & \textbf{83} & \textbf{21} & \textbf{25}  \\ \botline
\end{tabular}}
\end{table}

\begin{table}
\caption{\textit{Evaluation in the Gulf Stream region.} Our model trained specifically on the Gulf Stream region outperforms \hresvthree\ trained on the extratropical Global area, and also achieves higher accuracy than alternative methods. The boost of regional training is particularly significant for forecasts at time $T+7$. \hresvthree\ trained on the Global area achieves similar or slightly better performance than delayed-time methods DUACS and NeurOST, and better performance than operational Mercator forecasts. \metricdesc\ } 
\label{tab:gs}
\resizebox{\textwidth}{!}{\begin{tabular}{b{3.5cm}b{0.8cm}b{1.5cm}b{0.8cm}b{1.5cm}b{0.8cm}b{1.5cm}}
    \topline
    \metricLines
    DUACS (DT pers.)
    & 83 & 61 & 71 & 65 & 29 & 42\\ \midline
    NeurOST (DT pers.)
    & 86 & 63 & 72 & 65 & 25 & 37 \\ \midline
    Mercator
    & 66 & 52 & 65 & 64 & 38  & 41  \\ \midline
    \hresvthree~Global
    & 85 & 66 & 74 & 66 & 27 & 34 \\ \midline
    \hresvthree~Regional 
    & \textbf{88} & \textbf{73} & \textbf{77} & \textbf{70} & \textbf{24} & \textbf{31} \\ \botline
\end{tabular}}
\end{table}

In contrast, the Gulf Stream and Agulhas regions are highly energetic regions, presenting a unique challenge for forecasting. The regional models outperform the global model, particularly at 7-day lead times ($T+7$), showcasing the ability of regional models to capture local dynamics (Tables~\ref{tab:gs} and~\ref{tab:agh}). In the Gulf Stream region, the regional model achieves 88\% correct angle predictions, outperforming the global model at 85\% and surpassing the performance of delayed-time methods such as DUACS (83\%) and NeurOST (86\%), as well as near-real-time Mercator (77\%) (Table~\ref{tab:gs}). Furthermore, at $T+7$, the regional model predicts angles with 73\% accuracy, compared to 66\% for the global model. Similarly, in the Agulhas region, the regional model achieves an angle accuracy of 93\% at time $T+1$, compared to 92\% for the global model (Table~\ref{tab:agh}). At $T+7$, the regional model reaches 78\% accuracy in angle prediction, surpassing the global model's 73\% and indicating its superior ability to capture the region’s ocean dynamics over extended time frames. These results emphasise the advantage of training on region-specific data for accurate, long-term forecasts in highly energetic ocean regions.

\begin{table}
\caption{\textit{Evaluation in the Agulhas region.} Our model trained specifically on the Aghulas region outperforms \hresvthree\ trained on the extratropical Global area, and also achieves higher accuracy than alternative methods. The boost of regional training is particularly significant for forecasts at time $T+7$. \hresvthree\ trained on the Global area achieves better performance than delayed-time methods DUACS and NeurOST, as well as operational Mercator forecasts. \metricdesc\ } 
\label{tab:agh}
\resizebox{\textwidth}{!}{\begin{tabular}{b{3.7cm}b{0.8cm}b{1.5cm}b{0.8cm}b{1.5cm}b{0.8cm}b{1.5cm}}
    \topline
    \metricLines
    DUACS (DT pers.)
    & 87 & 67 & 71 & 67 & 29 & 35 \\ \midline
    NeurOST (DT pers.)
    & 91 & 70 & 75 & 67 & 26 & 35 \\ \midline
    Mercator &  77 & 62 & 64 & 66 & 37 & 38 \\ \midline
    \hresvthree~Global 
    & 92 & 73 & 77 & 70 & 25 & 33 \\ \midline
    \hresvthree~Regional 
    & \textbf{93} & \textbf{78} & \textbf{80} & \textbf{73} & \textbf{23} & \textbf{30} \\ \botline
\end{tabular}}
\end{table}

\begin{figure}
    \centering
    \includegraphics[width=0.9\linewidth]{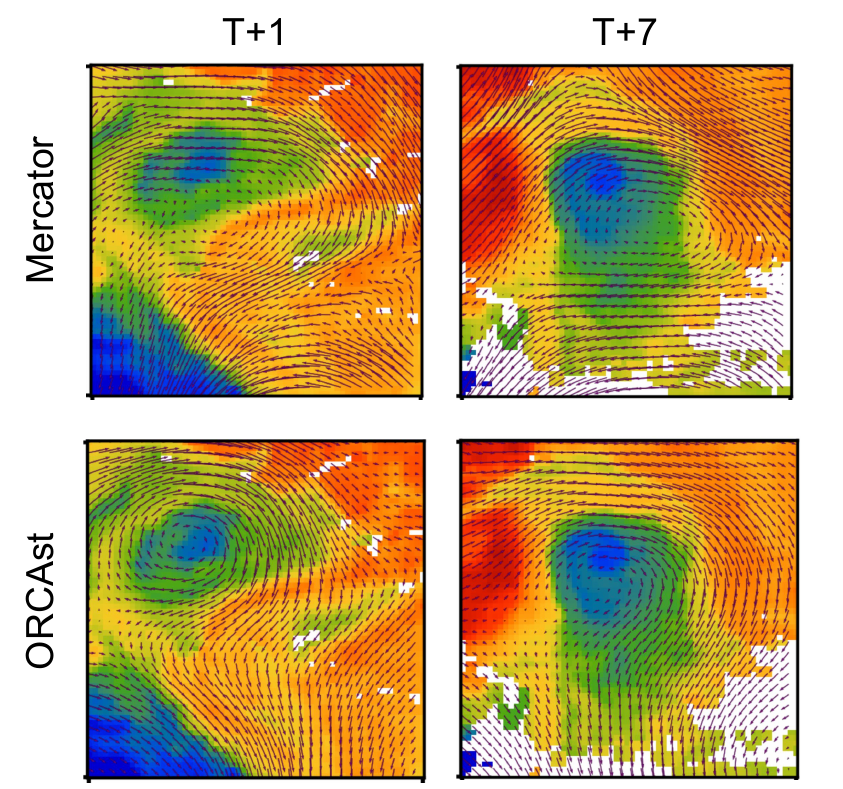}
    \caption{Predicted evolution of a structure in the Agulhas region between October 11 and October 17, 2023, within the latitude and longitude bounds of 39.9°S–37.8°S and 29.1°E–31.2°E. SST is shows in the background, with black arrows denoting the ocean surface currents. The Mercator model misidentifies the structure's location but predicts a westward movement. In contrast, \hresvthree\ both accurately locates the structure and provides a precise 7-day forecast of its future position. }
    \label{fig:quali_forecast}
\end{figure}

Figure~\ref{fig:quali_forecast} illustrates how \hresvthree\ learns to accurately predict the evolution of a dynamic structure in the Agulhas region. The Mercator model incorrectly localises the structure, but predicts a westward evolution. In contrast, \hresvthree\ not only correctly identifies the location of the structure, but also accurately forecasts its position in 7-days. This ability to both localise and predict the evolution of such features highlights the strength of \hresvthree\ in learning complex ocean dynamics.

\subsection{Including more input data}\label{results:inputdata}

The flexibility of \hresvthree\ allows us to include various inputs. In this section, we present experiments where we evaluate \hresvthree\ using CHL or SWOT data as inputs, in addition to SST and nadir altimetry. Table~\ref{tab:chl_inp} compares model performance using CHL as additional inputs, and Table~\ref{tab:swot_inp} compares model performance using SWOT as additional inputs. 

Table~\ref{tab:chl_inp} suggests that there is limited potential of CHL to provide complementary high-resolution information on ocean currents, when used in combination with high-resolution SST data. Scores for the model both with and without CHL are similar. 

\begin{table}
\caption{Comparison of ORCAst performances on the extratropical Global area, using SST or SST + CHL as inputs, in addition to nadir SSH. Including CHL data does not bring important improvements to \hresvthree~ currents. \metricdesc\ } 
\label{tab:chl_inp}
\resizebox{\textwidth}{!}{\begin{tabular}{b{3.7cm}b{0.8cm}b{1.5cm}b{0.8cm}b{1.5cm}b{0.8cm}b{1.5cm}}
    \topline
    \metricLines
    \hresvthree\ SST
    & \textbf{85} & \textbf{70} & \textbf{77} & 69 & \textbf{24} & 30 \\ \midline
    \hresvthree\ SST+CHL
    & \textbf{85} & \textbf{70} & 76 & \textbf{72} & 27 & \textbf{29}  \\ \botline
\end{tabular}}
\end{table}

We compare \hresvthree\ models using nadir-pointing altimetry only or nadir and SWOT as SSH inputs, in addition to SST. As SWOT data is only available from mid-2023 onwards, we train our model using SWOT inputs during Stage 2 and evaluate our model using August-December 2023. We keep the training produce of Stage 1 as before, using only SST and nadir altimetry as inputs. Results of this model at the end of Stage 2 are provided in Table~\ref{tab:swot_inp}. Using SWOT data as inputs seemingly does not help the model to better predict ocean surface currents. However, since there is only around one year of available SWOT data, the training and test sets are small, making it difficult to draw strong conclusions from this experiment. SWOT observations reveal unobserved mesoscale and sub-mesoscale structure, however the long revisit time and limited available data from the mission may partially explain the lack of performance gains when used as additional inputs.

\begin{table}
\caption{Comparison of ORCAst performances on the extratropical Global area, using nadir altimetry or SWOT and nadir altimetry as inputs, in addition to SST. Including SWOT data as inputs does not bring improvements to \hresvthree\ currents.  This model was trained as usual during Stage 1, then trained with SWOT inputs during Stage 2. It is evaluated on August-December 2023, rather than the whole of 2023, due to availability of SWOT data, thus the first line of results is not identical to the line corresponding to Stage 2 in Table~\ref{tab:stage_importance}. } 
\label{tab:swot_inp}
\resizebox{\textwidth}{!}{\begin{tabular}{b{3.9cm}b{0.8cm}b{1.5cm}b{0.8cm}b{1.5cm}b{0.8cm}b{1.5cm}}
    \topline
    \metricLines
    \hresvthree\ nadir-only
    & \textbf{83} & \textbf{68} & \textbf{73} & \textbf{69} & \textbf{25} & \textbf{30} \\ \midline
    \hresvthree\ nadir+SWOT
    & 82 & 67 & 72 & 68 & 26 & 31 \\ \botline 
\end{tabular}}
\end{table}

\subsection{Qualitative evaluation}\label{results:quali}

\begin{figure}
    \centering
    \includegraphics[width=\linewidth]{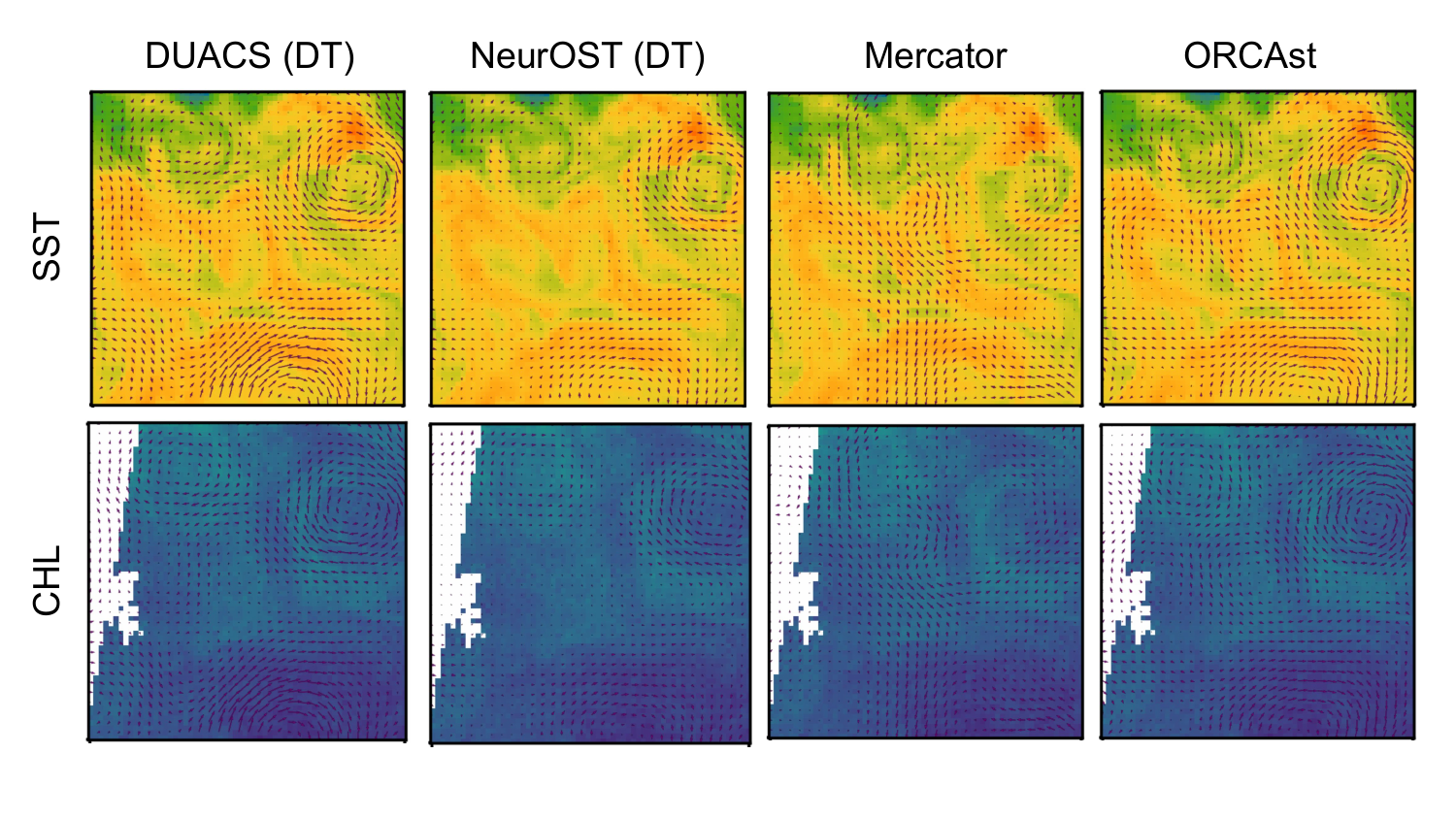}
    \caption{Model performance in the Mediterranean Sea on June 28, 2023, within the latitude and longitude bounds of 32.7°N–34.8°N and 24.6°W–26.7°W. The top row shows the corresponding SST images in the background, and the bottow row shows the corresponding CHL images in the background. The field of ocean surface currents for each model is represented using black arrows. \hresvthree\ correctly localises both large and fine-scale structures, visible in the SST and CHL signatures. We note that for this example, we illustrate a higher-resolution DUACS product, only available for the Mediterranean Sea~\citep{medduacs}.}
    \label{fig:quali_comparison}
\end{figure}

An example of \hresvthree\ predicted currents at time $T+1$ in the Mediterranean Sea is shown in Figure~\ref{fig:quali_comparison}, alongside comparisons with delayed-time models from DUACS and NeurOST, as well as ocean surface currents from Mercator. SST and CHL images are shown in the background. Although DUACS and NeurOST tend to correctly localise large-scale structures, some smaller-scale structures may be missed. In the case of DUACS, this is because the method lacks high-resolution information from SST. In the case of NeurOST, this is possibly because the model learns from nadir altimetry, with a low effective resolution, and is thus not sufficiently penalised for not learning fine-scale information visible in the SST. Figure~\ref{fig:quali_twostages} illustrates the difference between training on nadir altimetry and high-resolution SWOT altimtery. 

\begin{figure}
    \centering
    \includegraphics[width=\linewidth]{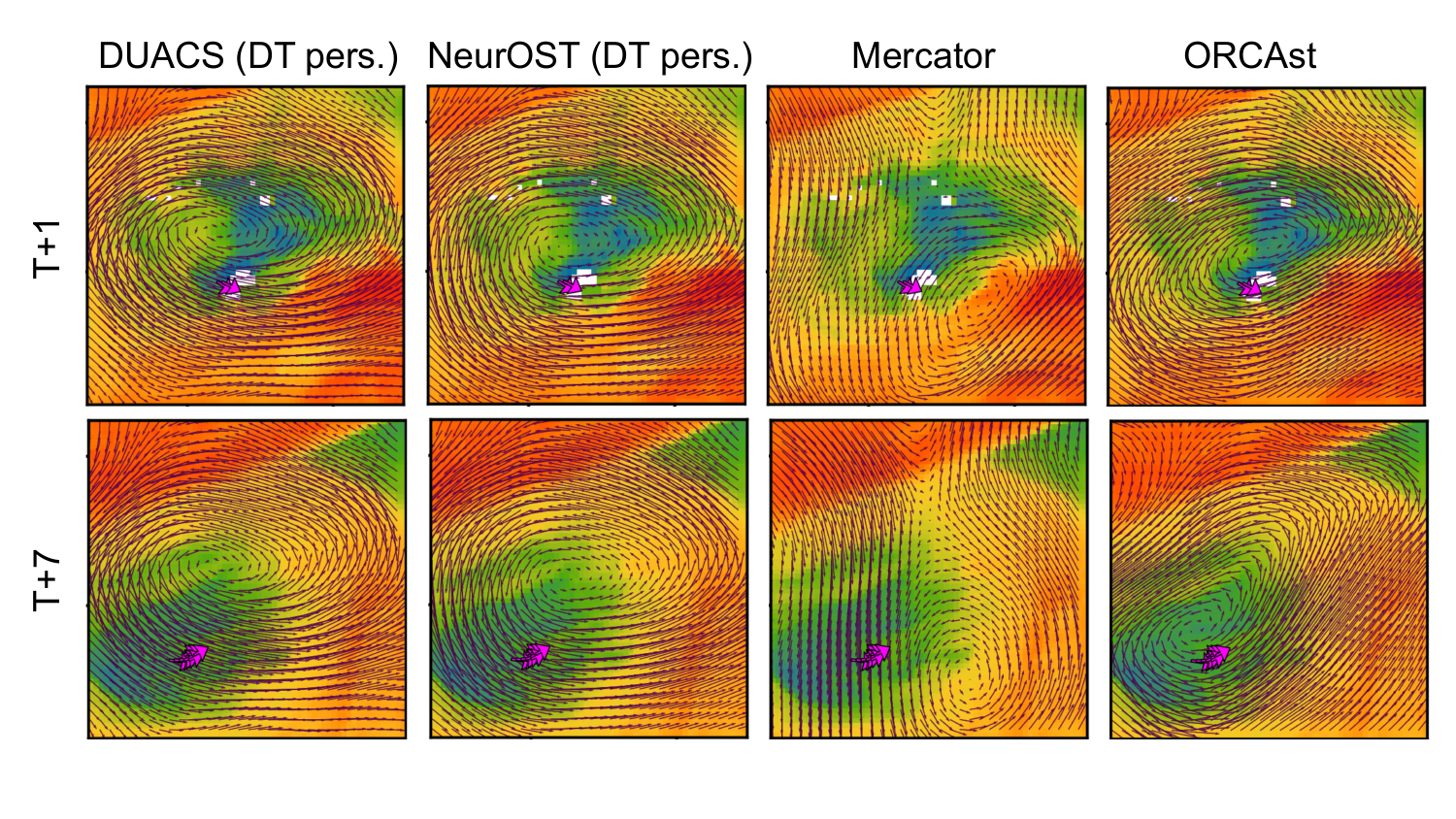}
    \caption{Drifter (in magenta) with WMO identification number 4402878 on August 12 and August 18, 2023, located within the latitude and longitude bounds of 37.8°N–39.9°N and 59.6°W–57.5°W. The image backgrounds show the corresponding SST, with the field of predicted ocean surface currents represented using black arrows. We compare the model performance to in situ observations from drifters at different lead times. \hresvthree\ correctly localises the eddy and predicts its evolution over 7 days.}
    \label{fig:quali_drifters}
\end{figure}

A qualitative comparison of our model outputs to drifter observations at days $T+1$ and $T+7$ is shown in Figure~\ref{fig:quali_drifters}. Like DUACS and NeurOST, \hresvthree\ correctly localises the eddy at $T+1$. However, persistence forecasts for DUACS and NeurOST do not align well with the drifter's motion at time $T+7$. In contrast, \hresvthree\ correctly predict the drifter’s trajectory at time $T+7$. Mercator fails to localise the eddy at $T+1$, as well as its evolution at time $T+7$.

\subsection{Evaluation using ship data}\label{results:ship}

Ships generally measure their Speed Through Water (STW) and Speed Over Ground (SOG), and the difference between these measurements, $\mathrm{SOG} - \mathrm{STW}$, serves as an indicator of the currents experienced by ships along their trajectory. 

\begin{figure}[htbp]
    \centering
    \includegraphics[width=0.95\textwidth]{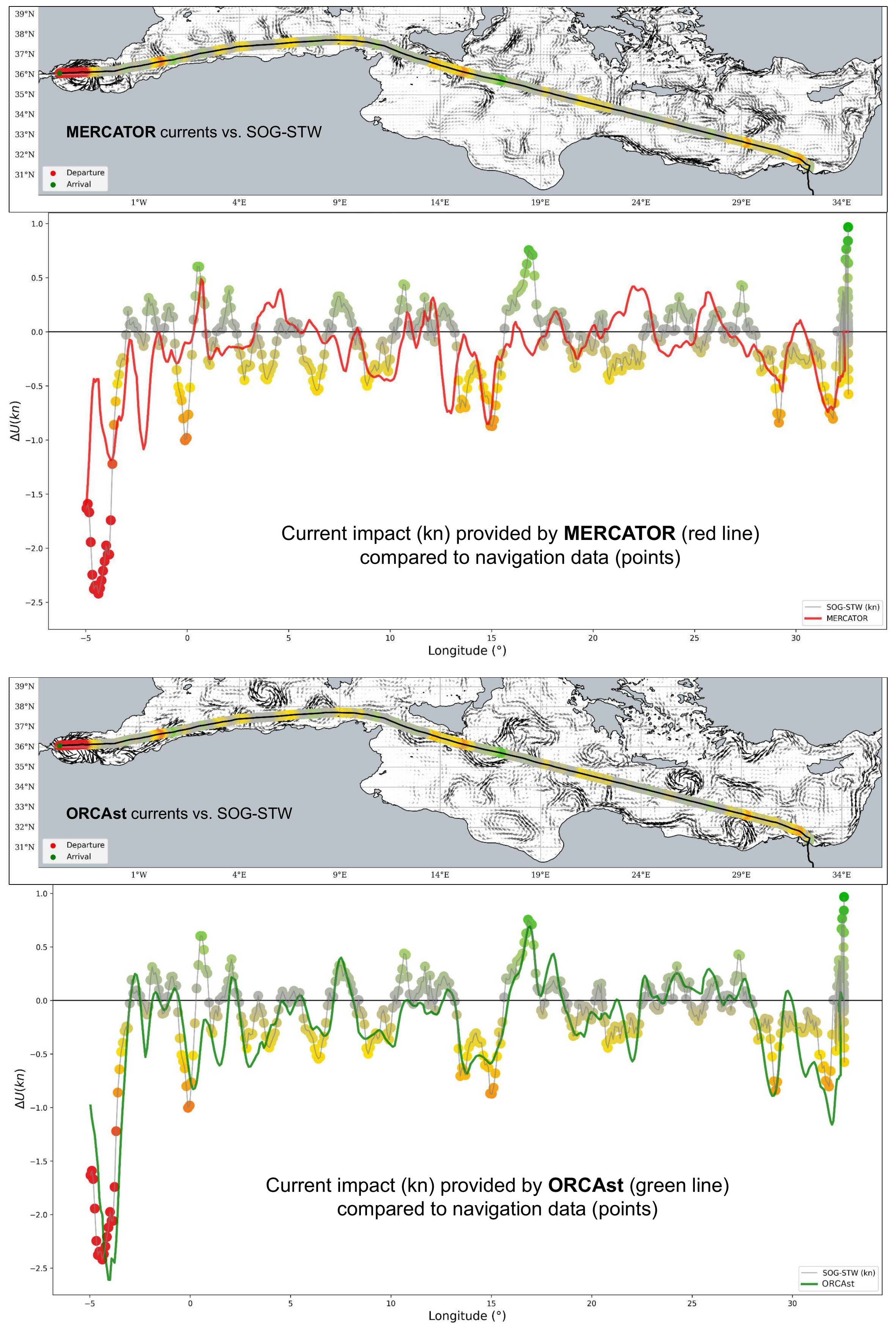}
    \caption{Evaluation of Mercator and ORCAst on ship data for a trans-Mediterranean voyage on August 8, 2023. Maps show ocean currents estimated by Mercator and ORCAst, with the ship route coloured proportionally to SOG-STW measurements. The graphs show SOG-STW measurements, compared with the current impact predicted by Mercator and ORCAst. The SOG-STW curve is coloured proportionally to SOG-STW, in the same way as in the map. Currents predicted by ORCAst more closely follow the SOG-STW measurements than currents predicted by Mercator. }
    \label{fig:ship_quali}
\end{figure}

In Figure~\ref{fig:ship_quali}, we evaluate our model using ship data from a trans-Mediterranean voyage on August 21, 2023. We compute $\mathrm{SOG} - \mathrm{STW}$ and compare these observations to predicted ocean currents along the ship's trajectory, as provided by the ORCAst and Mercator models. Our findings indicate that ORCAst predictions align more closely with the observed $\mathrm{SOG} - \mathrm{STW}$ values than Mercator, suggesting superior performance. Correct estimations of the currents felt along ship routes is important for maritime route optimisation, allowing ships to take routes following more favourable ocean currents and avoiding currents against the path of the ship. 

\section{Conclusion}\label{sec:discussion}

\subsection{Discussion}

In this study, we introduce \hresvthree, a deep learning framework for forecasting ocean surface currents. \hresvthree\ produces 1/30° high-resolution 7-day forecasts of ocean currents in Global extratropical regions. Our flexible approach allows us to fuse multivariate observations of the ocean such as in situ measurements of currents, satellite altimetry, and satellite imaging of sea surface temperature and chlorophyll-a, to improve the spatial resolution and prediction of the temporal evolution of surface currents. 

For next-day predictions, we accurately predict the angle of drifters 85\% of the time, surpassing delayed-time DUACS (78\%), delayed-time NeurOST (83\%) and Mercator (70\%). For 7-day forecasts, we achieve 70\% correct angle predictions, surpassing persistence forecasts of DUACS (62\%) and NeurOST (65\%), and numerical forecasts of the operational OGCM Mercator (56\%). We also achieve superior performance on our metrics of correct magnitude and mean MEVA. The superior performance of \hresvthree\ over delayed-time (DUACS and NeurOST) and near real-time (Mercator) methods highlights the potential of deep learning in leveraging large quantities of observational data to produce accurate operational ocean current forecasts.

We show that progressively training the network on sparser, but more direct, observations of currents, leads to improved performances at each training stage. We train using nadir-pointing altimetry and DUACS geostrophic currents, followed by SWOT SSH and SWOT geostrophic currents, and finally with drifter current measurements. In particular, training on drifters allows the model to learn from observations beyond geostrophic ocean surface currents. 

Regional models consistently outperform the globally trained model. The Mediterranean Sea, with its unique dynamics and fine spatial structures, showed the largest improvement when using a regional model. Our regional models also outperform the Global model in energetic regions such as the Gulf Stream and Agulhas regions, better capturing localised physical dynamics over longer lead times.

The inclusion of chlorophyll-a data did not yield substantial improvements when SST data was already used, suggesting limited complementary information from chlorophyll-a for forecasting surface currents. Similarly, incorporating SWOT data as inputs did not significantly enhance model performance, likely due to the limited availability of SWOT observations (one year of data). The addition of SWOT measures over the coming years should further enhance ORCAst's learning and performance. However, as there is little overlap between the SWOT measurements in 2023 and the drifter trajectories used for validation we cannot demonstrate yet the effectiveness of using SWOT data as inputs. We note that training our model using SWOT data as targets led to significant improvements for surface currents at meso and sub-mesoscale. 

The qualitative assessments reinforce the quantitative metrics, demonstrating that \hresvthree\ can provide detailed forecasts that align closely with observed patterns, outperforming baseline methods. Moreover, the evaluation against real ship data highlights the practical applicability of the model. Accurate current predictions are essential for optimizing ship routing and reducing emissions, making \hresvthree\ a valuable tool for the maritime industry.

\subsection{Perspectives}
Throughout this study, we demonstrate the potential of fusing multivariate observations with deep learning. However, a number of issues still need to be addressed to improve the quality of our forecasted ocean currents. 

First, \hresvthree~ produces no forecast between latitudes -20° and 20°. This is because the geostrophic approximation is not valid near the Equator, and the link between currents and altimetry weakens. One alternative approach would be to train a deep learning model using an assimilated numerical model as a target on the Equator. 

Second, to further develop the fusion of observations from various sources, we are interested in using ship data. For instance, the Automatic Identification System (AIS), provides information about the ship's steering and speed over ground, which can help us deduce the surface water velocity. However, using AIS data to obtain indirect current measurements is difficult, and calls for additional studies. Other sources of on-board observations exist, such as Acoustic Doppler Current Profiler (ADCP) observations or Wavex sensor data. These more direct measurement of currents along the ship's route can also be used as targets for evaluation or training, however they may also be noisy, often measuring incoherent current velocities.

Finally, from a methodological point of view, we are interested in going beyond regression models (like \hresvthree). Regression models estimate the mean of a distribution, their predictions are therefore smoothed, leading to a domain gap compared to real physical samples, with smoothing getting stronger when conditions are scarce or irrelevant. On the other hand, generative deep learning allows the network to generate multiple examples from the same set of inputs, hopefully forming a physics-matching distribution. In particular, Denoising Diffusion Probabilistic Models (DDPM) and flow matching techniques have led to an impressive image generation breakthrough in the past years. Using these sampling models, we might be able to estimate the full conditional distribution of the currents, instead of only the mean forecast.

\begin{ack}
This research study was supported by a Cifre thesis grant from the ANRT 2023/1590, and partially financed by the iLab2023 grant DOS0220247/00 of the French Public Bank of Investment (BPIFrance). This work was granted access to the HPC resources of IDRIS under the allocation AD011015571 made by GENCI. The authors of this paper have affiliations with Amphitrite, a company developing maritime solutions such as optimal ship routing, using methods similar to those discussed in this study. As such, there exists a potential conflict of interest regarding the presentation and interpretation of the results. The authors declare that the study was conducted with rigor and integrity, and the interpretation of the results is based on the evidence presented in the data.
\end{ack}

\counterwithin{figure}{section}
\counterwithin{table}{section}

\section*{Appendix A: Learned Positional Embeddings}\label{apx:pos_enc}
\renewcommand{\thetable}{A.\arabic{table}}
\renewcommand{\thefigure}{A.\arabic{figure}}
\setcounter{table}{0} %
\setcounter{figure}{0} %

\begin{figure}[!h]
    \centering
    \includegraphics[width=\linewidth]{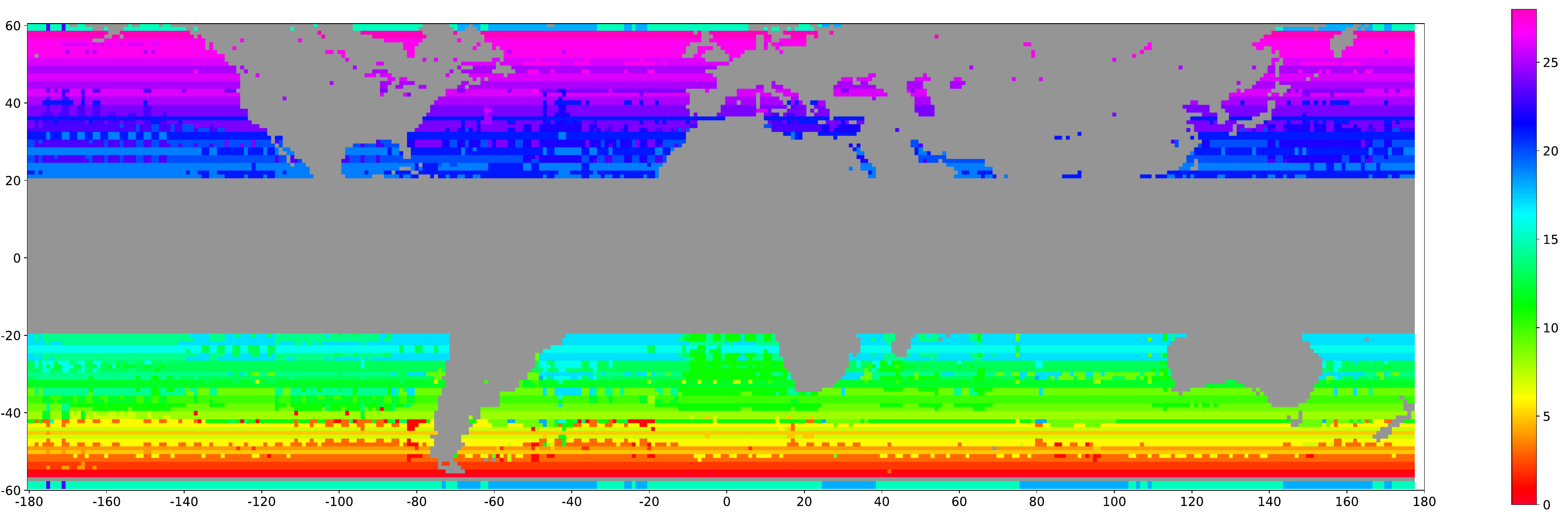}
    \caption{Clustering of positional embeddings using $k$-means, where $k=30$. Each colour represents a cluster learned by our model. We reorder the numbering and colours of the groups for clearer visualisation purposes. Land areas and the equator are in grey.}
    \label{fig:emb_clust}
\end{figure}

Learned positional embeddings provide insights into the spatio-temporal conditions most relevant for \hresvthree\ to forecast ocean currents. These embeddings are represented as 32-dimensional vectors. To visualise patterns within the embeddings, we create a 3D latitude-longitude-time grid encompassing all spatio-temporal positions used during inference. To efficiently cluster points in this large grid, we apply the mini-batch $k$-means clustering algorithm with $k=30$ clusters and a batch size of 8. Figure~\ref{fig:emb_clust} plots these $k=30$ clusters on a latitude-longitude grid for a chosen temporal dimension (first week of the year). Latitude is a strong determining factor of cluster membership, demonstrating that the positional embedding learns the strong influence of latitude in predicting ocean currents. 

\begin{figure}[!h]
    \centering
    \includegraphics[trim=0cm 9cm 0cm 0cm, clip, width=0.98\linewidth]{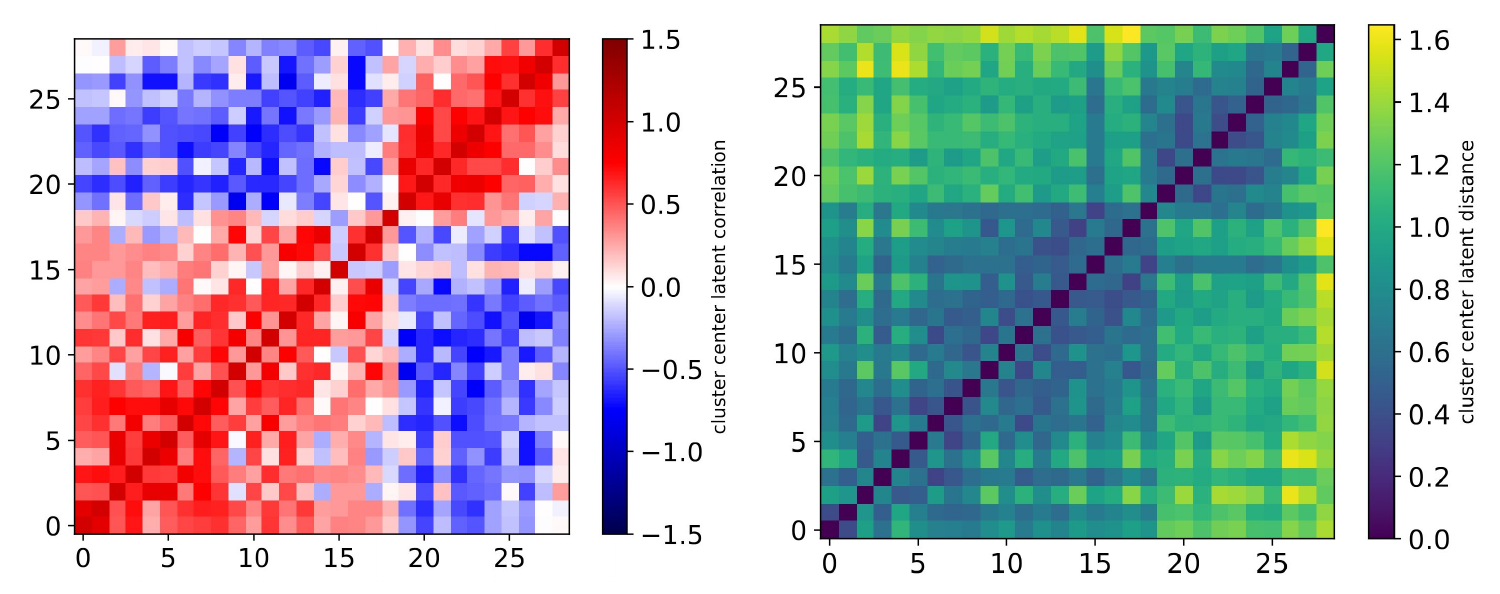}
    \caption{The correlation (left) and the distance (right) between clusters of our positional embedding, chosen using the $k$-means method with $k=30$. Geographic zones corresponding to cluster numbers are shows in Figure~\ref{fig:emb_clust}. Southern Hemisphere clusters are numbered 0-18 and Northern Hemisphere clusters are numbered 19-29. }
    \label{fig:emb_clust_combined}
\end{figure}

Furthermore, Figures~\ref{fig:emb_clust_combined} shows the correlation and the distance between clusters. 
We see a clear separation between Southern Hemisphere (clusters 0 to 18) and Northern Hemisphere (clusters 19-29) and Southern Hemisphere embeddings. The Southern Hemisphere is more represented in the clustering (19 clusters versus 11 clusters), because the ocean-to-continent ratio is higher in the South, which means that the model sees more South examples, and adapts its encoding accordingly.
Generally, Southern Hemisphere clusters are correlated with each other, and negatively correlated with Northern Hemisphere clusters. Southern Hemisphere clusters are closest to other Southern Hemisphere clusters, and Northern Hemisphere clusters are closest to other Northern Hemisphere clusters.

This is expected, as the Coriolis effect varies with latitude, reaching its maximum at the poles. In the Northern Hemisphere, the westerlies drive currents eastward, while the Coriolis effect deflects them to the right, forming clockwise gyres. In the Southern Hemisphere, a similar mechanism creates counterclockwise gyres due to the Coriolis effect deflecting currents to the left. The variability in the Coriolis effect modulates the intensity and structure of ocean currents in both hemispheres. 

\section*{Appendix B: Bias of SWOT}\label{apx:karin_nadir_bias}
\renewcommand{\thetable}{B.\arabic{table}}
\renewcommand{\thefigure}{B.\arabic{figure}}
\setcounter{table}{0} %
\setcounter{figure}{0} %

\begin{figure}[!ht]
    \centering
    \includegraphics[width=1.\textwidth]{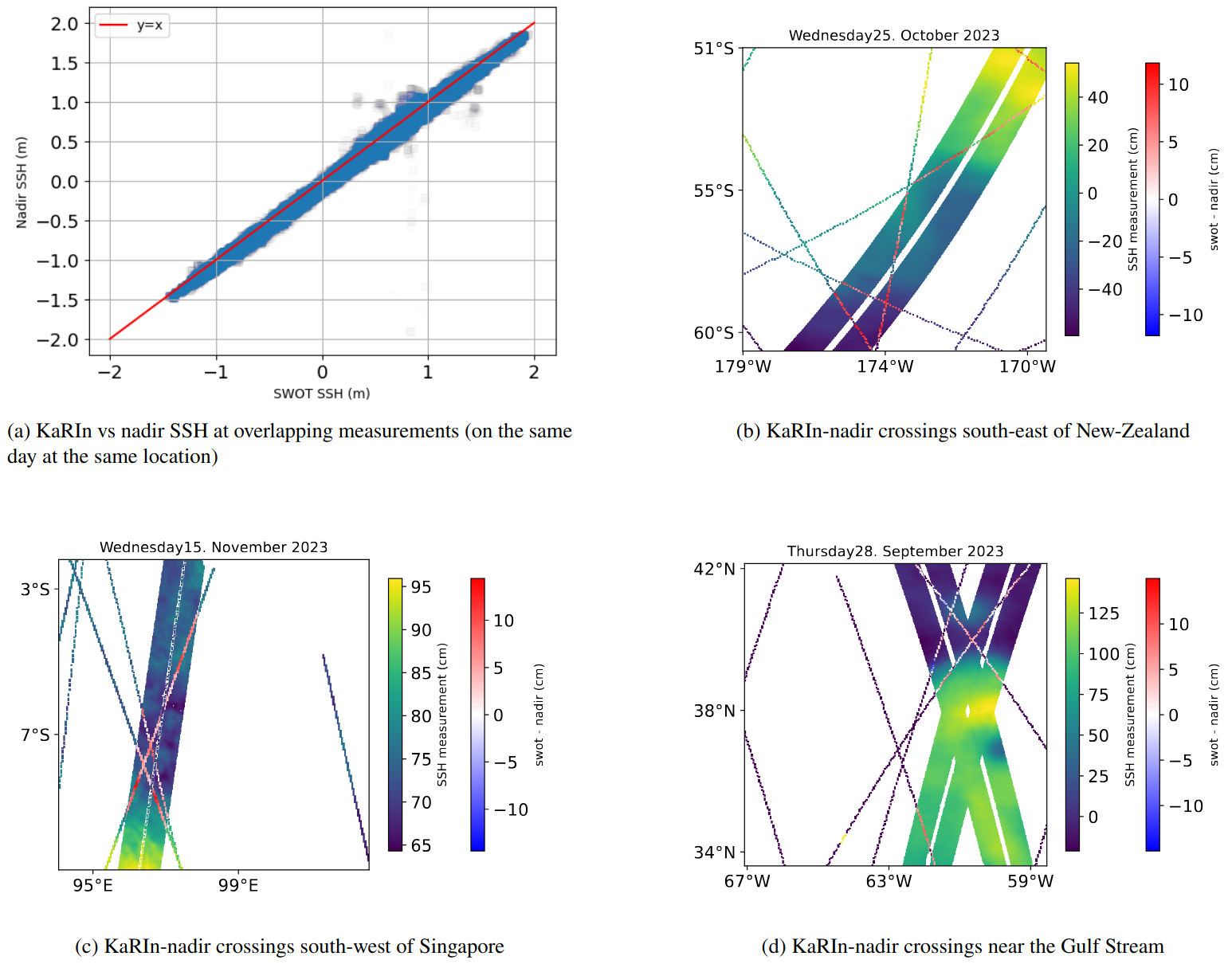}
    \caption{We compare KaRIn and nadir altimetry across intersection points of both measures of SSH. We find that there is an underestimation of SSH from nadir compared to SSH measured by SWOT's KaRIn instrument, of around $5.26 \pm 3.32$\unit{\centi\meter}. }
    \label{fig:karinXNadir}
\end{figure}

We calculate the bias between ADT measured by SWOT and ADT measured by nadir altimetry at the same location and on the same day, as well as the standard deviation of the differences (Figure~\ref{fig:karinXNadir}). We find that SWOT's KaRIn instrument overestimates SSH compared to nadir altimetry by $5.26 \pm 3.32$~\unit{\centi\meter}. This bias may make training more complicated for our model, trained using nadir SSH in Stage 1 and SWOT SSH in Stage 2. Correcting this bias could help our model better learn SSH from both nadir and SWOT's KaRIn instrument. As our main focus is learning ocean surface currents and not absolute values of SSH, we suppose that this bias is less important for our model.

\section*{Appendix C: Ablation Study}\label{apx:neurostAblation}
\renewcommand{\thetable}{C.\arabic{table}}
\renewcommand{\thefigure}{C.\arabic{figure}}
\setcounter{table}{0} %
\setcounter{figure}{0} %

NeurOST performs better than DUACS, as shown in Table~\ref{tab:global_comp}, however we train using DUACS as targets in Stage 1 of our learning prodecure. As shown in Table~\ref{tab:neurostVsDuacs}, replacing DUACS with NeurOST improves model performance during Stage 1 of training. However, the difference in performance difference diminishes in Stages 2 and 3. 

\begin{table}
\caption{Comparison of model performance using DUACS and NeurOST as targets during Stage 1 of training in the Gulf Stream region. Replacing DUACS with NeurOST improves performance in Stage 1 of training. However, the performance difference diminishes in Stages 2 and 3, where the models achieve similar results. \metricdesc\ }
    \label{tab:neurostVsDuacs}
    \resizebox{\textwidth}{!}{\begin{tabular}{b{5.0cm}b{0.8cm}b{1.5cm}b{0.8cm}b{1.5cm}b{0.8cm}b{1.5cm}}
        \topline
        \metricLines
        NeurOST (DT pers.)
        & 86 & 63 & 72 & 65 & 25 & 37\\ \midline
        Stage 1
        & 82 & 67 & 67 & 65 & 29 & 33 \\ 
        Stage 1 on NeurOST
        & \textbf{87} & \textbf{72} & \textbf{73} & \textbf{68} & \textbf{26} & \textbf{32} \\ \midline
        Stage 2
        & 86 & \textbf{72} & \textbf{76} & 72 & \textbf{25}  & 32 \\ 
        Stage 2, NeurOST for Stage 1
        & \textbf{87} & \textbf{72} & \textbf{76} & \textbf{71} & \textbf{25} & \textbf{31} \\ \midline
        Stage 3
        & 88 & \textbf{73} & 77 & 70 & 24 & 31 \\ 
        Stage 3, NeurOST for Stage 1
        & \textbf{89} & \textbf{73} & \textbf{79} & \textbf{71} & \textbf{23} & \textbf{30} \\ \botline
    \end{tabular}}
\end{table}

\clearpage

\begingroup
\sloppy

\endgroup


\begin{thebibliography}{43}
\providecommand{\natexlab}[1]{#1}
\providecommand{\url}[1]{\texttt{#1}}
\expandafter\ifx\csname urlstyle\endcsname\relax
  \providecommand{\doi}[1]{doi: #1}\else
  \providecommand{\doi}{doi: \begingroup \urlstyle{rm}\Url}\fi

\bibitem[glo()]{globalsst}
{GHRSST} {NOAA}/{STAR} {ACSPO} v2.81 0.02 degree {L3S} {Daily} {Dataset} from {LEO} {Satellites}.
\newblock URL \url{https://podaac.jpl.nasa.gov/dataset/L3S_LEO_DY-STAR-v2.81}.

\bibitem[Aouni et~al.(2024)Aouni, Gaudel, Regnier, Van~Gennip, Drevillon, Drillet, and Lellouche]{aouni2024glonet}
A.~E. Aouni, Q.~Gaudel, C.~Regnier, S.~Van~Gennip, M.~Drevillon, Y.~Drillet, and J.-M. Lellouche.
\newblock Glonet: Mercator's end-to-end neural forecasting system.
\newblock \emph{arXiv preprint arXiv:2412.05454}, 2024.

\bibitem[Archambault et~al.(2023)Archambault, Filoche, Charantonis, and B{\'e}r{\'e}ziat]{archambault2023visapp}
T.~Archambault, A.~Filoche, A.~Charantonis, and D.~B{\'e}r{\'e}ziat.
\newblock Multimodal unsupervised spatio-temporal interpolation of satellite ocean altimetry maps.
\newblock In \emph{International Conference on Computer Vision Theory and Applications (VISAPP)}, Lisboa, Portugal, Feb. 2023.

\bibitem[Archambault et~al.(2024{\natexlab{a}})Archambault, Filoche, Charantonis, and Béréziat]{archambault_pre-training_2024}
T.~Archambault, A.~Filoche, A.~Charantonis, and D.~Béréziat.
\newblock Pre-training and {Fine}-tuning {Attention} {Based} {Encoder} {Decoder} {Improves} {Sea} {Surface} {Height} {Multi}-variate {Inpainting}.
\newblock In \emph{{VISAPP} 2024 - 19th {International} {Conference} on {Computer} {Vision} {Theory} and {Applications}}, Roma, Italy, Feb. 2024{\natexlab{a}}.
\newblock URL \url{https://hal.sorbonne-universite.fr/hal-04475205}.

\bibitem[Archambault et~al.(2024{\natexlab{b}})Archambault, Filoche, Charantonis, Béréziat, and Thiria]{archambault_learning_2024}
T.~Archambault, A.~Filoche, A.~A. Charantonis, D.~Béréziat, and S.~Thiria.
\newblock Learning {Sea} {Surface} {Height} {Interpolation} from {Multi}-variate {Simulated} {Satellite} {Observations}.
\newblock \emph{James, Journal of Advancing in Modeling Earth Syst}, 16\penalty0 (6):\penalty0 e2023MS004047, 2024{\natexlab{b}}.
\newblock \doi{10.1029/2023ms004047}.
\newblock URL \url{https://hal.sorbonne-universite.fr/hal-04568666}.
\newblock Publisher: AGU.

\bibitem[Callaham et~al.(2019)Callaham, Maeda, and Brunton]{callaham_robust_2019}
J.~L. Callaham, K.~Maeda, and S.~L. Brunton.
\newblock Robust flow reconstruction from limited measurements via sparse representation.
\newblock \emph{Physical Review Fluids}, 4\penalty0 (10):\penalty0 103907, Oct. 2019.
\newblock \doi{10.1103/PhysRevFluids.4.103907}.
\newblock URL \url{https://link.aps.org/doi/10.1103/PhysRevFluids.4.103907}.
\newblock Publisher: American Physical Society.

\bibitem[Chen et~al.(2023)Chen, Zhong, Zhang, Cheng, Xu, Qi, and Li]{chen_fuxi_2023}
L.~Chen, X.~Zhong, F.~Zhang, Y.~Cheng, Y.~Xu, Y.~Qi, and H.~Li.
\newblock {FuXi}: a cascade machine learning forecasting system for 15-day global weather forecast.
\newblock \emph{npj Climate and Atmospheric Science}, 6\penalty0 (1):\penalty0 1--11, Nov. 2023.
\newblock ISSN 2397-3722.
\newblock \doi{10.1038/s41612-023-00512-1}.
\newblock URL \url{https://www.nature.com/articles/s41612-023-00512-1}.
\newblock Publisher: Nature Publishing Group.

\bibitem[Ciani et~al.(2024)Ciani, Fanelli, and Buongiorno~Nardelli]{ciani_estimating_2024}
D.~Ciani, C.~Fanelli, and B.~Buongiorno~Nardelli.
\newblock Estimating ocean currents from the joint reconstruction of absolute dynamic topography and sea surface temperature through deep learning algorithms.
\newblock \emph{EGUsphere}, pages 1--25, Apr. 2024.
\newblock \doi{10.5194/egusphere-2024-1164}.
\newblock URL \url{https://egusphere.copernicus.org/preprints/2024/egusphere-2024-1164/}.
\newblock Publisher: Copernicus GmbH.

\bibitem[{Copernicus Climate Service}(2024)]{globalduacs}
{Copernicus Climate Service}.
\newblock {Global Ocean Gridded L4 Sea Surface Heights And Derived Variables Reprocessed 1993 Ongoing}.
\newblock {Satellite observations, NetCDF-4 format}, 11 2024.
\newblock URL \url{https://doi.org/10.48670/moi-00148}.
\newblock Product ID: SEALEVEL\_GLO\_PHY\_L4\_MY\_008\_047.

\bibitem[{Copernicus Marine Service}(2023)]{sshtracks}
{Copernicus Marine Service}.
\newblock {Global Ocean Along Track L3 Sea Surface Heights NRT}.
\newblock {Satellite observations, NetCDF-4 format}, 11 2023.
\newblock URL \url{https://doi.org/10.48670/moi-00147}.
\newblock Product ID: SEALEVEL\_GLO\_PHY\_L3\_NRT\_008\_044.

\bibitem[{Copernicus Marine Service}(2024)]{medduacs}
{Copernicus Marine Service}.
\newblock {European Seas Gridded L4 Sea Surface Heights And Derived Variables NRT}.
\newblock {Satellite observations, NetCDF-4 format}, 11 2024.
\newblock URL \url{https://doi.org/10.48670/moi-00142}.
\newblock Product ID: SEALEVEL\_EUR\_PHY\_L4\_NRT\_008\_060.

\bibitem[Drévillon et~al.(2008)Drévillon, Bourdallé-Badie, Derval, Drillet, Lellouche, Rémy, Tranchant, Benkiran, Greiner, Guinehut, Verbrugge, Garric, Testut, Laborie, Nouel, Bahurel, Bricaud, Crosnier, Dombrowsky, Durand, Ferry, Hernandez, Galloudec, Messal, and Parent]{drevillon_godaemercator-ocean_2008}
M.~Drévillon, R.~Bourdallé-Badie, C.~Derval, Y.~Drillet, J.-M. Lellouche, E.~Rémy, B.~Tranchant, M.~Benkiran, E.~Greiner, S.~Guinehut, N.~Verbrugge, G.~Garric, C.-E. Testut, M.~Laborie, L.~Nouel, P.~Bahurel, C.~Bricaud, L.~Crosnier, E.~Dombrowsky, E.~Durand, N.~Ferry, F.~Hernandez, O.~L. Galloudec, F.~Messal, and L.~Parent.
\newblock The {GODAE}/{Mercator}-{Ocean} global ocean forecasting system: results, applications and prospects.
\newblock \emph{Journal of Operational Oceanography}, 1\penalty0 (1):\penalty0 51--57, Feb. 2008.

\bibitem[Elipot et~al.(2016)Elipot, Lumpkin, Perez, Lilly, Early, and Sykulski]{elipot2016globaldrifter}
S.~Elipot, R.~Lumpkin, R.~C. Perez, J.~M. Lilly, J.~J. Early, and A.~M. Sykulski.
\newblock {A global surface drifter dataset at hourly resolution}.
\newblock \emph{{Journal of Geophysical Research: Oceans}}, 121, 2016.
\newblock \doi{10.1002/2016JC011716}.

\bibitem[Elipot et~al.(2022)Elipot, Sykulski, Lumpkin, Centurioni, and Pazos]{driftersgdp}
S.~Elipot, A.~Sykulski, R.~Lumpkin, L.~Centurioni, and M.~Pazos.
\newblock {Hourly location, current velocity, and temperature collected from Global Drifter Program drifters world-wide}.
\newblock Dataset, 2022.
\newblock URL \url{https://doi.org/10.25921/x46c-3620}.
\newblock {Accessed on July 31, 2023}.

\bibitem[Emery et~al.(1989)Emery, Brown, and Nowak]{emery_avhrr_1989}
W.~Emery, J.~Brown, and Z.~Nowak.
\newblock {AVHRR} image navigation - {Summary} and review.
\newblock \emph{Photogrammetric Engineering \& Remote Sensing}, 55, Sept. 1989.

\bibitem[{EUMETSAT for Copernicus}(2024)]{chldata}
{EUMETSAT for Copernicus}.
\newblock {Ocean Colour Service: Product Data Format Specification - OLCI Level 1 \& Level 2 Instrument Products}, 2024.
\newblock URL \url{https://data.eumetsat.int/product/EO:EUM:DAT:0407}.
\newblock Platform: Sentinel-3; Sensor: OLCI (Optical); Data Level: Level 2; Collection ID: EO:EUM:DAT:0407; Parameters: Ocean, Ocean Colour, Level 2 Data.

\bibitem[Fablet et~al.(2024)Fablet, Chapron, Le~Sommer, and Sévellec]{fablet_inversion_2024}
R.~Fablet, B.~Chapron, J.~Le~Sommer, and F.~Sévellec.
\newblock Inversion of {Sea} {Surface} {Currents} {From} {Satellite}-{Derived} {SST}-{SSH} {Synergies} {With} {4DVarNets}.
\newblock \emph{Journal of Advances in Modeling Earth Systems}, 16\penalty0 (6):\penalty0 e2023MS003609, 2024.
\newblock ISSN 1942-2466.
\newblock \doi{10.1029/2023MS003609}.
\newblock URL \url{https://onlinelibrary.wiley.com/doi/abs/10.1029/2023MS003609}.
\newblock \_eprint: https://onlinelibrary.wiley.com/doi/pdf/10.1029/2023MS003609.

\bibitem[Filoche et~al.(2022)Filoche, Archambault, Charantonis, and Béréziat]{filoche2022}
A.~Filoche, T.~Archambault, A.~Charantonis, and D.~Béréziat.
\newblock Statistics-free interpolation of ocean observations with deep spatio-temporal prior.
\newblock In \emph{{ECML/PKDD Workshop on Machine Learning for Earth Observation and Prediction (MACLEAN)}}, 2022.

\bibitem[Gao et~al.(2022)Gao, Tan, Wu, and Li]{gao_simvp_2022}
Z.~Gao, C.~Tan, L.~Wu, and S.~Z. Li.
\newblock Simvp: Simpler yet better video prediction.
\newblock In \emph{2022 IEEE/CVF Conference on Computer Vision and Pattern Recognition (CVPR)}, pages 3160--3170, 2022.
\newblock \doi{10.1109/CVPR52688.2022.00317}.

\bibitem[Gurvan et~al.(2017)Gurvan, Bourdallé-Badie, Bouttier, Bricaud, Bruciaferri, Calvert, Chanut, Clementi, Coward, Delrosso, Ethé, Flavoni, Graham, Harle, Iovino, Lea, Lévy, Lovato, Martin, Masson, Mocavero, Paul, Rousset, Storkey, Storto, and Vancoppenolle]{gurvan_nemo_2017}
M.~Gurvan, R.~Bourdallé-Badie, P.-A. Bouttier, C.~Bricaud, D.~Bruciaferri, D.~Calvert, J.~Chanut, E.~Clementi, A.~Coward, D.~Delrosso, C.~Ethé, S.~Flavoni, T.~Graham, J.~Harle, D.~Iovino, D.~Lea, C.~Lévy, T.~Lovato, N.~Martin, S.~Masson, S.~Mocavero, J.~Paul, C.~Rousset, D.~Storkey, A.~Storto, and M.~Vancoppenolle.
\newblock {NEMO} ocean engine, Oct. 2017.
\newblock URL \url{https://zenodo.org/records/3248739}.
\newblock Publisher: Zenodo.

\bibitem[{Ifremer}(2024)]{drifterscmems}
{Ifremer}.
\newblock {Global Ocean In-Situ Near-Real-Time Observations}.
\newblock {In-situ observations, NetCDF-4 format}, 11 2024.
\newblock URL \url{https://doi.org/10.48670/moi-00036}.
\newblock Product ID: INSITU\_GLO\_PHYBGCWAV\_DISCRETE\_MYNRT\_013\_030.

\bibitem[Kingma and Ba(2014)]{kingma2014adam}
D.~P. Kingma and J.~Ba.
\newblock Adam: A method for stochastic optimization.
\newblock In \emph{International Conference on Learning Representations (ICLR)}, 2014.

\bibitem[Kugusheva et~al.(2024)Kugusheva, Bull, Moschos, Ioannou, Le~Vu, and Stegner]{kugusheva_ocean_2024}
A.~Kugusheva, H.~Bull, E.~Moschos, A.~Ioannou, B.~Le~Vu, and A.~Stegner.
\newblock Ocean {Satellite} {Data} {Fusion} for {High}-{Resolution} {Surface} {Current} {Maps}.
\newblock \emph{Remote Sensing}, 16\penalty0 (7):\penalty0 1182, Mar. 2024.
\newblock ISSN 2072-4292.
\newblock \doi{10.3390/rs16071182}.
\newblock URL \url{https://www.mdpi.com/2072-4292/16/7/1182}.

\bibitem[Lam et~al.(2023)Lam, Sanchez-Gonzalez, Willson, Wirnsberger, Fortunato, Alet, Ravuri, Ewalds, Eaton-Rosen, Hu, Merose, Hoyer, Holland, Vinyals, Stott, Pritzel, Mohamed, and Battaglia]{lam_learning_2023}
R.~Lam, A.~Sanchez-Gonzalez, M.~Willson, P.~Wirnsberger, M.~Fortunato, F.~Alet, S.~Ravuri, T.~Ewalds, Z.~Eaton-Rosen, W.~Hu, A.~Merose, S.~Hoyer, G.~Holland, O.~Vinyals, J.~Stott, A.~Pritzel, S.~Mohamed, and P.~Battaglia.
\newblock Learning skillful medium-range global weather forecasting.
\newblock \emph{Science (New York, N.Y.)}, page eadi2336, Nov. 2023.
\newblock ISSN 1095-9203.
\newblock \doi{10.1126/science.adi2336}.

\bibitem[Le~Guillou et~al.(2023{\natexlab{a}})Le~Guillou, Gaultier, Ballarotta, Metref, Ubelmann, Cosme, and Rio]{le_guillou_regional_2023}
F.~Le~Guillou, L.~Gaultier, M.~Ballarotta, S.~Metref, C.~Ubelmann, E.~Cosme, and M.-H. Rio.
\newblock Regional mapping of energetic short mesoscale ocean dynamics from altimetry: performances from real observations.
\newblock \emph{Ocean Science}, 19\penalty0 (5):\penalty0 1517--1527, Oct. 2023{\natexlab{a}}.
\newblock ISSN 1812-0784.
\newblock \doi{10.5194/os-19-1517-2023}.
\newblock URL \url{https://os.copernicus.org/articles/19/1517/2023/}.
\newblock Publisher: Copernicus GmbH.

\bibitem[Le~Guillou et~al.(2023{\natexlab{b}})Le~Guillou, Maxime, Metref, Ubelman, Cosme, and M.-H.]{ODC}
F.~Le~Guillou, B.~Maxime, S.~Metref, C.~Ubelman, E.~Cosme, and R.~M.-H.
\newblock The woc data challenges.
\newblock \url{https://2024-dc-woc-esa.readthedocs.io/en/latest/index.html}, 2023{\natexlab{b}}.
\newblock Regional mapping of energetic short mesoscale ocean dynamics from altimetry: performances from real observations.

\bibitem[Martin et~al.(2023)Martin, Manucharyan, and Klein]{martin_synthesizing_2023}
S.~A. Martin, G.~E. Manucharyan, and P.~Klein.
\newblock Synthesizing {Sea} {Surface} {Temperature} and {Satellite} {Altimetry} {Observations} {Using} {Deep} {Learning} {Improves} the {Accuracy} and {Resolution} of {Gridded} {Sea} {Surface} {Height} {Anomalies}.
\newblock \emph{Journal of Advances in Modeling Earth Systems}, 15\penalty0 (5):\penalty0 e2022MS003589, 2023.
\newblock ISSN 1942-2466.
\newblock \doi{10.1029/2022MS003589}.
\newblock URL \url{https://onlinelibrary.wiley.com/doi/abs/10.1029/2022MS003589}.
\newblock \_eprint: https://onlinelibrary.wiley.com/doi/pdf/10.1029/2022MS003589.

\bibitem[Martin et~al.(2024)Martin, Manucharyan, and Klein]{martin_deep_2024}
S.~A. Martin, G.~E. Manucharyan, and P.~Klein.
\newblock Deep {Learning} {Improves} {Global} {Satellite} {Observations} of {Ocean} {Eddy} {Dynamics}.
\newblock \emph{Geophysical Research Letters}, 51\penalty0 (17):\penalty0 e2024GL110059, 2024.
\newblock ISSN 1944-8007.
\newblock \doi{10.1029/2024GL110059}.
\newblock URL \url{https://onlinelibrary.wiley.com/doi/abs/10.1029/2024GL110059}.
\newblock \_eprint: https://onlinelibrary.wiley.com/doi/pdf/10.1029/2024GL110059.

\bibitem[{Mercator Océan International}(2024)]{mercatordata}
{Mercator Océan International}.
\newblock {Global Ocean Physics Analysis and Forecast}, 2024.
\newblock URL \url{https://doi.org/10.48670/moi-00016}.
\newblock Product ID: GLOBAL\_ANALYSISFORECAST\_PHY\_001\_024.

\bibitem[{MET Norway}(2023)]{medsst}
{MET Norway}.
\newblock {Mediterranean Sea - High Resolution and Ultra High Resolution L3S Sea Surface Temperature}, 2023.
\newblock URL \url{https://doi.org/10.48670/moi-00171}.
\newblock Processing Level: 3; Spatial Coverage: Mediterranean Sea (Lat 30.25° to 46°, Lon -18.12° to 36.25°); Spatial Resolution: 0.01° × 0.01°; Temporal Coverage: 1 Jan 2008 to 17 Dec 2024; Update Frequency: Daily; Format: NetCDF-4.

\bibitem[Morrow et~al.(2018)Morrow, Blurmstein, and Dibarboure]{morrow2018fine}
R.~Morrow, D.~Blurmstein, and G.~Dibarboure.
\newblock Fine-scale altimetry and the future swot mission.
\newblock \emph{New frontiers in operational oceanography}, pages 191--226, 2018.

\bibitem[NASA/JPL and CNES(2024)]{DATAswot}
NASA/JPL and CNES.
\newblock {The SWOT\_L3\_LR\_SSH product, derived from the {L2} {SWOT} KaRIn low rate ocean data products [Dataset]}, 2024.
\newblock \url{https://doi.org/10.24400/527896/A01-2023.018}.

\bibitem[O'Reilly et~al.(1998)O'Reilly, Maritorena, Mitchell, Siegel, Carder, Garver, Kahru, and Mcclain]{oreilly_ocean_1998}
J.~O'Reilly, S.~Maritorena, B.~Mitchell, D.~Siegel, K.~Carder, S.~Garver, M.~Kahru, and C.~Mcclain.
\newblock Ocean color chlorophyll algorithms for {SeaWiFS}.
\newblock \emph{Journal of Geophysical Research}, 103:\penalty0 937--953, Oct. 1998.

\bibitem[Scott et~al.(2024)Scott, Georgy, and Patrice]{scott_neurost}
A.~M. Scott, E.~M. Georgy, and K.~Patrice.
\newblock {NeurOST Level 4 Sea Surface Height and Surface Geostrophic Currents Analysis Product (Version 2024.0)}.
\newblock {Daily mapped neural network product using Level 3 altimetry observations and MUR Level 4 SST}, 08 2024.
\newblock URL \url{https://doi.org/10.5067/NEURO-STV24}.
\newblock Start/Stop Date: 2010-Jan-01 to 2024-Jun-15; Format: netCDF-4; Processing Level: 4.

\bibitem[Taburet et~al.(2019)Taburet, Sanchez-Roman, Ballarotta, Pujol, Legeais, Fournier, Faugere, and Dibarboure]{taburet_duacs_2019}
G.~Taburet, A.~Sanchez-Roman, M.~Ballarotta, M.-I. Pujol, J.-F. Legeais, F.~Fournier, Y.~Faugere, and G.~Dibarboure.
\newblock {DUACS} {DT2018}: 25 years of reprocessed sea level altimetry products.
\newblock \emph{Ocean Science}, 15\penalty0 (5):\penalty0 1207--1224, Sept. 2019.
\newblock ISSN 1812-0784.
\newblock \doi{10.5194/os-15-1207-2019}.
\newblock URL \url{https://os.copernicus.org/articles/15/1207/2019/}.
\newblock Publisher: Copernicus GmbH.

\bibitem[Tapley et~al.(1982)Tapley, Born, and Parke]{tapley1982seasat}
B.~D. Tapley, G.~H. Born, and M.~E. Parke.
\newblock The seasat altimeter data and its accuracy assessment.
\newblock \emph{Journal of Geophysical Research: Oceans}, 87\penalty0 (C5):\penalty0 3179--3188, 1982.

\bibitem[Thiria et~al.(2023)Thiria, Sorror, Archambault, Charantonis, Bereziat, Mejia, Molines, and Crépon]{thiria_downscaling_2023}
S.~Thiria, C.~Sorror, T.~Archambault, A.~Charantonis, D.~Bereziat, C.~Mejia, J.-M. Molines, and M.~Crépon.
\newblock Downscaling of ocean fields by fusion of heterogeneous observations using {Deep} {Learning} algorithms.
\newblock \emph{Ocean Modelling}, 182:\penalty0 102174, Apr. 2023.
\newblock ISSN 1463-5003.
\newblock \doi{10.1016/j.ocemod.2023.102174}.
\newblock URL \url{https://www.sciencedirect.com/science/article/pii/S146350032300015X}.

\bibitem[Tonani et~al.(2015)Tonani, Balmaseda, Bertino, Blockley, Brassington, Davidson, Drillet, Hogan, Kuragano, Lee, Mehra, Paranathara, Tanajura, and Wang]{tonani_status_2015}
M.~Tonani, M.~Balmaseda, L.~Bertino, E.~Blockley, G.~Brassington, F.~Davidson, Y.~Drillet, P.~Hogan, T.~Kuragano, T.~Lee, A.~Mehra, F.~Paranathara, C.~A. Tanajura, and H.~Wang.
\newblock Status and future of global and regional ocean prediction systems.
\newblock \emph{Journal of Operational Oceanography}, 8\penalty0 (sup2):\penalty0 s201--s220, Aug. 2015.
\newblock ISSN 1755-876X.
\newblock \doi{10.1080/1755876X.2015.1049892}.
\newblock URL \url{https://doi.org/10.1080/1755876X.2015.1049892}.
\newblock Publisher: Taylor \& Francis \_eprint: https://doi.org/10.1080/1755876X.2015.1049892.

\bibitem[Ubelmann et~al.(2022)Ubelmann, Carrere, Durand, Dibarboure, Faugère, Ballarotta, Briol, and Lyard]{ubelmann_simultaneous_2022}
C.~Ubelmann, L.~Carrere, C.~Durand, G.~Dibarboure, Y.~Faugère, M.~Ballarotta, F.~Briol, and F.~Lyard.
\newblock Simultaneous estimation of ocean mesoscale and coherent internal tide sea surface height signatures from the global altimetry record.
\newblock \emph{Ocean Science}, 18\penalty0 (2):\penalty0 469--481, Apr. 2022.
\newblock ISSN 1812-0784.
\newblock \doi{10.5194/os-18-469-2022}.
\newblock URL \url{https://os.copernicus.org/articles/18/469/2022/}.
\newblock Publisher: Copernicus GmbH.

\bibitem[Wang et~al.(2022)Wang, Xie, Li, Fan, Song, Liang, Lu, Luo, and Shao]{wang_pvt_2022}
W.~Wang, E.~Xie, X.~Li, D.-P. Fan, K.~Song, D.~Liang, T.~Lu, P.~Luo, and L.~Shao.
\newblock {PVT} v2: {Improved} baselines with {Pyramid} {Vision} {Transformer}.
\newblock \emph{Computational Visual Media}, 8\penalty0 (3):\penalty0 415--424, Sept. 2022.
\newblock ISSN 2096-0662.
\newblock \doi{10.1007/s41095-022-0274-8}.
\newblock URL \url{https://doi.org/10.1007/s41095-022-0274-8}.

\bibitem[Wang et~al.(2024)Wang, Wang, Hu, Wang, Huo, Wang, Wang, Wang, Zhu, Xu, Yin, Bao, Luo, Zu, Han, Zhang, Ren, Deng, and Song]{wang_xihe_2024}
X.~Wang, R.~Wang, N.~Hu, P.~Wang, P.~Huo, G.~Wang, H.~Wang, S.~Wang, J.~Zhu, J.~Xu, J.~Yin, S.~Bao, C.~Luo, Z.~Zu, Y.~Han, W.~Zhang, K.~Ren, K.~Deng, and J.~Song.
\newblock {XiHe}: {A} {Data}-{Driven} {Model} for {Global} {Ocean} {Eddy}-{Resolving} {Forecasting}, Feb. 2024.
\newblock URL \url{http://arxiv.org/abs/2402.02995}.
\newblock arXiv:2402.02995 [physics].

\bibitem[Wu and He(2020)]{wu_group_2020}
Y.~Wu and K.~He.
\newblock Group {Normalization}.
\newblock \emph{International Journal of Computer Vision}, 128\penalty0 (3):\penalty0 742--755, Mar. 2020.
\newblock ISSN 1573-1405.
\newblock \doi{10.1007/s11263-019-01198-w}.
\newblock URL \url{https://doi.org/10.1007/s11263-019-01198-w}.

\bibitem[Yu et~al.(2022)Yu, Luo, Zhou, Si, Zhou, Wang, Feng, and Yan]{metaFormer}
W.~Yu, M.~Luo, P.~Zhou, C.~Si, Y.~Zhou, X.~Wang, J.~Feng, and S.~Yan.
\newblock Metaformer is actually what you need for vision.
\newblock In \emph{2022 IEEE/CVF Conference on Computer Vision and Pattern Recognition (CVPR)}, pages 10809--10819, 2022.
\newblock \doi{10.1109/CVPR52688.2022.01055}.

\end{thebibliography}
\end{document}